\documentclass[]{bytedance_seed}

\usepackage[toc,page,header]{appendix}

\usepackage{minitoc}

\title{Reinforcing Chain-of-Thought Reasoning with Self-Evolving Rubrics}

\author[1,2,*,\ddagger]{Leheng~Sheng}
\author[1,*,\dagger]{Wenchang~Ma}
\author[1]{Ruixin~Hong} 
\author[3]{\linebreak[4] Xiang~Wang}
\author[3,\dagger]{An~Zhang}
\author[2]{Tat-Seng~Chua}

\affiliation[1]{Seed, ByteDance}
\affiliation[2]{National University of Singapore}
\affiliation[3]{University~of~Science~and~Technology~of~China}

\contribution[*]{Equal Contribution}
\contribution[\ddagger]{Work done at ByteDance Seed}
\contribution[\dagger]{Corresponding authors}

\abstract{
Despite chain-of-thought (CoT) playing crucial roles in LLM reasoning, directly rewarding it is difficult: training a reward model demands heavy human labeling efforts, and static RMs struggle with evolving CoT distributions and reward hacking.
These challenges motivate us to seek an autonomous CoT rewarding approach that requires no human annotation efforts and can evolve gradually. 
Inspired by recent self-evolving training methods, we propose \textbf{RLCER} (\textbf{R}einforcement \textbf{L}earning with \textbf{C}oT Supervision via Self-\textbf{E}volving \textbf{R}ubrics), which enhances the outcome-centric RLVR by rewarding CoTs with self-proposed and self-evolving rubrics. 
We show that self-proposed and self-evolving rubrics provide reliable CoT supervision signals even without outcome rewards, enabling RLCER to outperform outcome-centric RLVR.
Moreover, when used as in-prompt hints, these self-proposed rubrics further improve inference-time performance.
}

\date{\today}

\correspondence{
An Zhang at \email{an\_zhang@ustc.edu.cn}
Wenchang Ma at \email{mawenchang.99@bytedance.com}
}

\checkdata[Project Page]{\url{https://alphalab-ustc.github.io/rlcer-alphalab/}}

\usepackage{amsmath}
\usepackage{amssymb}
\usepackage{mathtools}
\usepackage{amsthm}
\usepackage{soul}
\usepackage{wrapfig}

\theoremstyle{plain}

\theoremstyle{definition}

\theoremstyle{remark}

\usepackage[textsize=tiny]{todonotes}

\usepackage{enumitem}
\usepackage[table]{xcolor}  
\usepackage{amsthm}     
\usepackage{thmtools}   
\usepackage{tcolorbox}
\usepackage{tabularx}
\usepackage{multirow} 
\usepackage{dblfloatfix}

\usepackage{tikz}
\usetikzlibrary{calc,tikzmark,backgrounds}

\pgfdeclarelayer{algobg}
\pgfsetlayers{algobg,main}

\newcommand{\tikzmk}[1]{%
  \tikz[remember picture,overlay]\node[inner sep=0pt,outer sep=0pt] (#1) {};%
}

\newlength{\algoboxshiftx}
\newlength{\algoboxpadxL}
\newlength{\algoboxpadxR}
\newlength{\algoboxpadyT}
\newlength{\algoboxpadyB}

\definecolor{RolloutBg}{RGB}{234, 208, 228}
\definecolor{RewardBg}{RGB}{250, 235, 206}
\definecolor{PolicyBg}{RGB}{217, 215, 219}

\newcommand{\boxit}[4][]{%
\begin{tikzpicture}[remember picture,overlay]
\begin{pgfonlayer}{algobg}
  \coordinate (ALG@NW) at ([xshift=\algoboxshiftx-\algoboxpadxL,yshift=\algoboxpadyT]#3);

  \coordinate (ALG@SE) at ([xshift=\algoboxshiftx+\dimexpr\linewidth+\algoboxpadxR\relax,
                            yshift=-\algoboxpadyB]#3 |- #4);

  \fill[#2, opacity=0.14, rounded corners=2pt] (ALG@NW) rectangle (ALG@SE);

  \if\relax\detokenize{#1}\relax\else
    \node[anchor=south east, font=\scriptsize\bfseries, text=black!70]
      at ([xshift=-0.2em,yshift=0.2ex]ALG@SE) {#1};
  \fi
\end{pgfonlayer}
\end{tikzpicture}%
}

\newcommand{\ie}{\emph{i.e., }}
\newcommand{\eg}{\emph{e.g., }}

\newcommand{\cf}{\emph{cf. }}

\newcommand{\concept}[1]{{\color{black}{#1}}}
\newcommand{\revision}[1]{{\color{black}{#1}}}

\definecolor{lightgray}{gray}{0.95}  
\definecolor{table_color}{RGB}{252, 223, 208}

\usepackage{algorithm}
\usepackage{algorithmic}

\newcommand{\CommentBelow}[1]{%
  {\color{gray}\footnotesize // #1}%
}

\definecolor{good}{RGB}{58,113,104}
\definecolor{bad}{RGB}{180,0,0}
\definecolor{malboxborder}{RGB}{180,0,0}
\definecolor{benboxborder}{RGB}{81,91,131}
\definecolor{malboxbg}{RGB}{255,250,250}
\definecolor{benboxbg}{RGB}{251,252,254}
\definecolor{titlebg}{RGB}{77,77,77}
\definecolor{rubricator}{RGB}{220, 105, 123}
\definecolor{reasoner}{RGB}{22, 78, 164}

\definecolor{performance_color}{RGB}{255, 248, 249}
\definecolor{latency_color}{RGB}{246, 249, 254}

\newsavebox{\CaseStudy}
\newsavebox{\CaseStudyApdxOne}
\newsavebox{\CaseStudyApdxTwo}
\newsavebox{\CaseStudyApdxThree}
\newsavebox{\DSRPrompt}
\newsavebox{\CRPrompt}
\newsavebox{\GPTPrompt}

\newtcolorbox{gptpromptbox}[1][]{%
  colback=white,
  colframe=black,
  boxrule=1pt,
  arc=2pt,
  width=\textwidth,
  boxsep=2pt,
  left=2pt,
  right=2pt,
  top=2pt,
  bottom=2pt,
  title={PROMPT:}, 
  fonttitle=\large\bfseries, 
  coltitle=white, 
  colbacktitle=titlebg, 
  #1
}

\newtcolorbox{promptbox}[1][]{%
  colback=white,
  colframe=black,
  boxrule=1pt,
  sharp corners,
  width=\textwidth,
  boxsep=2pt,
  left=2pt,
  right=2pt,
  top=2pt,
  bottom=2pt,
  #1
}

\begin{document}
\maketitle


\begin{figure}[!h]
    \centering
    \begin{subfigure}[t]{0.605\columnwidth}
        \centering
        \includegraphics[width=\linewidth]{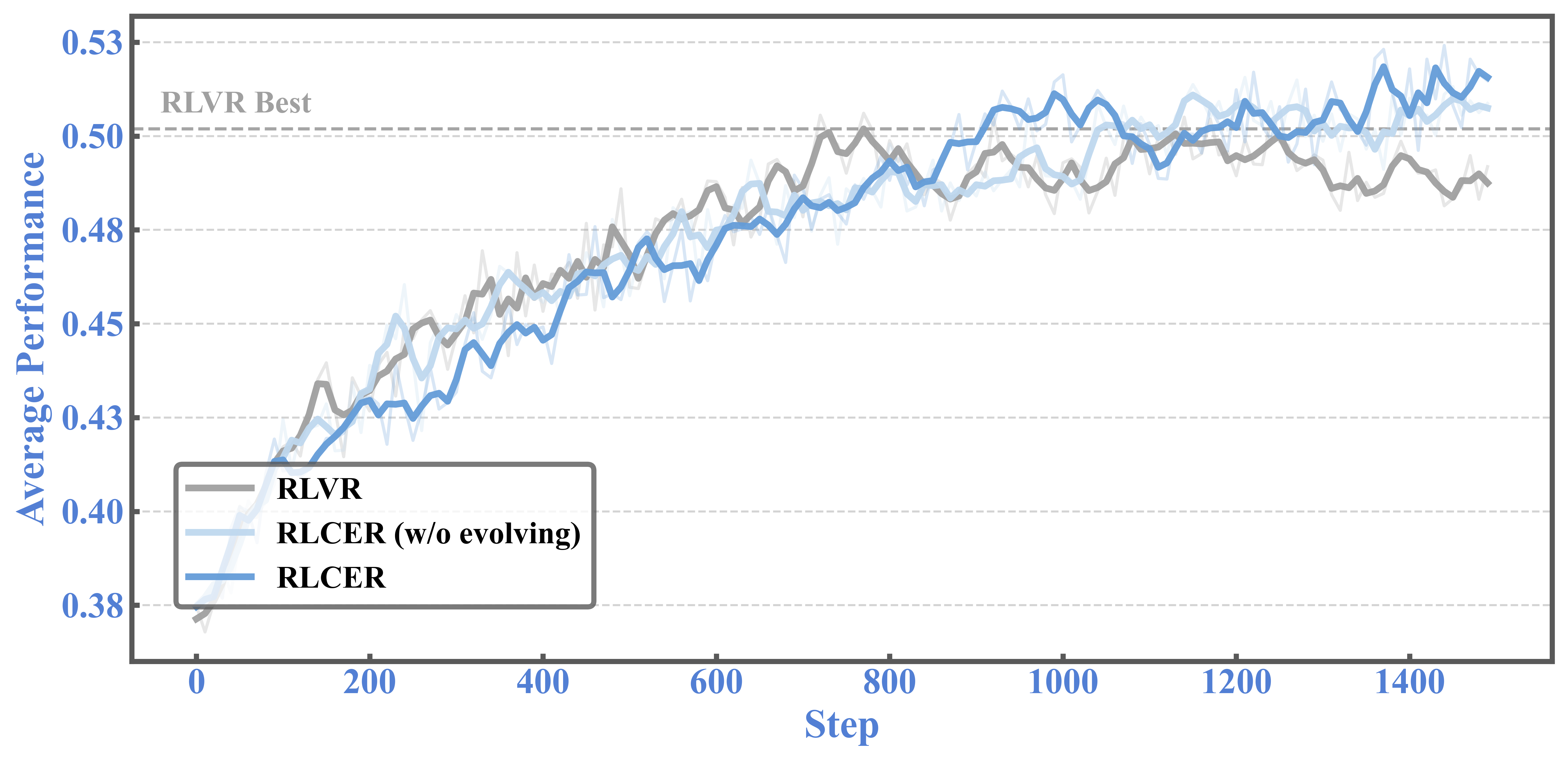}
        \vspace{-20pt}
        \caption{Average performance dynamics.}
        \label{fig:ablation-perf-dynamics}
    \end{subfigure}
    \begin{subfigure}[t]{0.382\columnwidth}
        \centering
        \includegraphics[width=\linewidth]{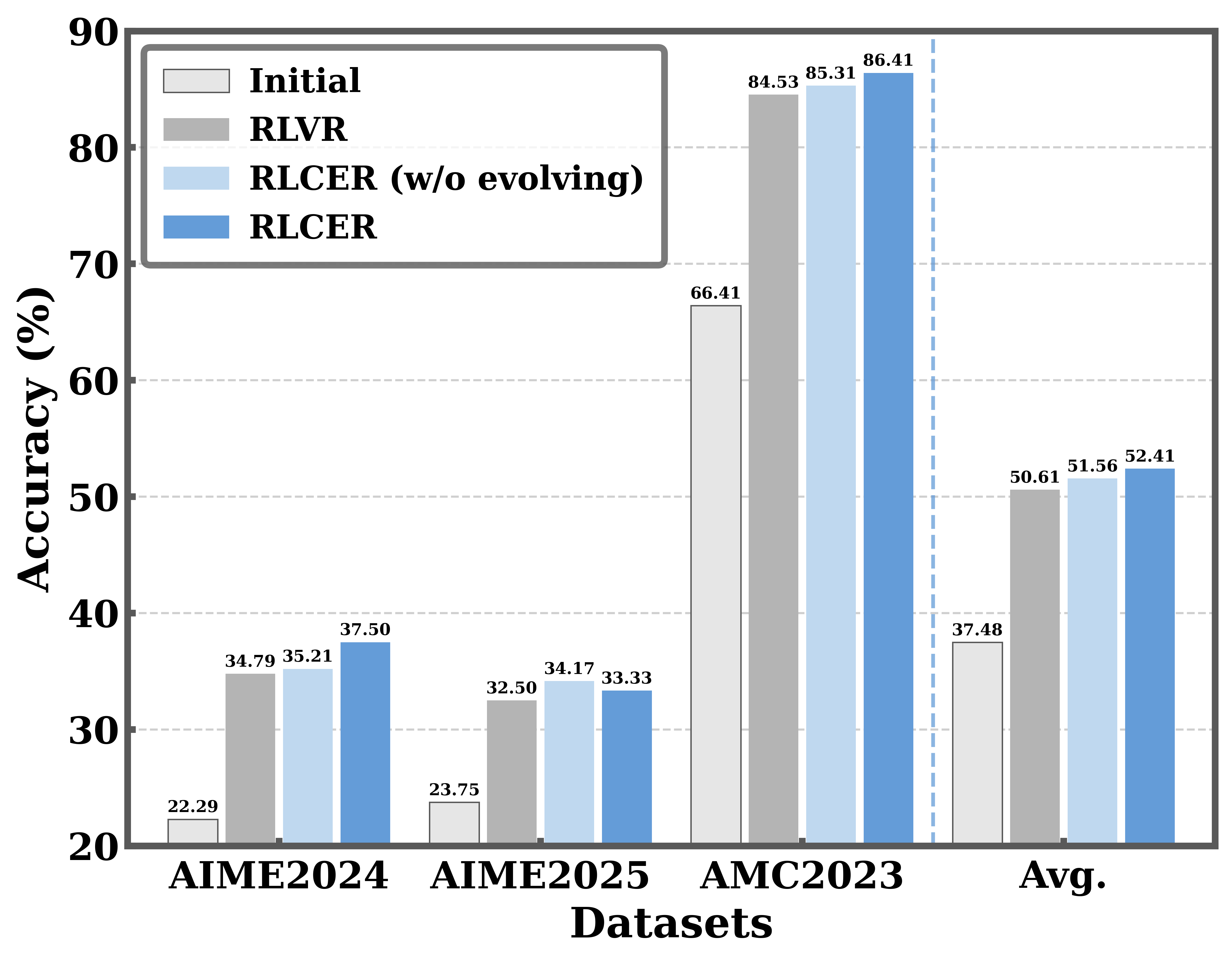} 
        \vspace{-20pt}
        \caption{Performance comparison.}
        \label{fig:ablation-perf-final}
    \end{subfigure}
    \vspace{-8pt}
    \caption{Performance across three math datasets on the 7B model. Training with RLCER leads to a higher performance ceiling, and the self-evolving rubrics further enhance reasoning performance. }
    \label{fig:ablation-perf}
\end{figure}

\clearpage
\section{Introduction}

\begin{figure}[ht]
    \centering
    \includegraphics[width=0.9\columnwidth]{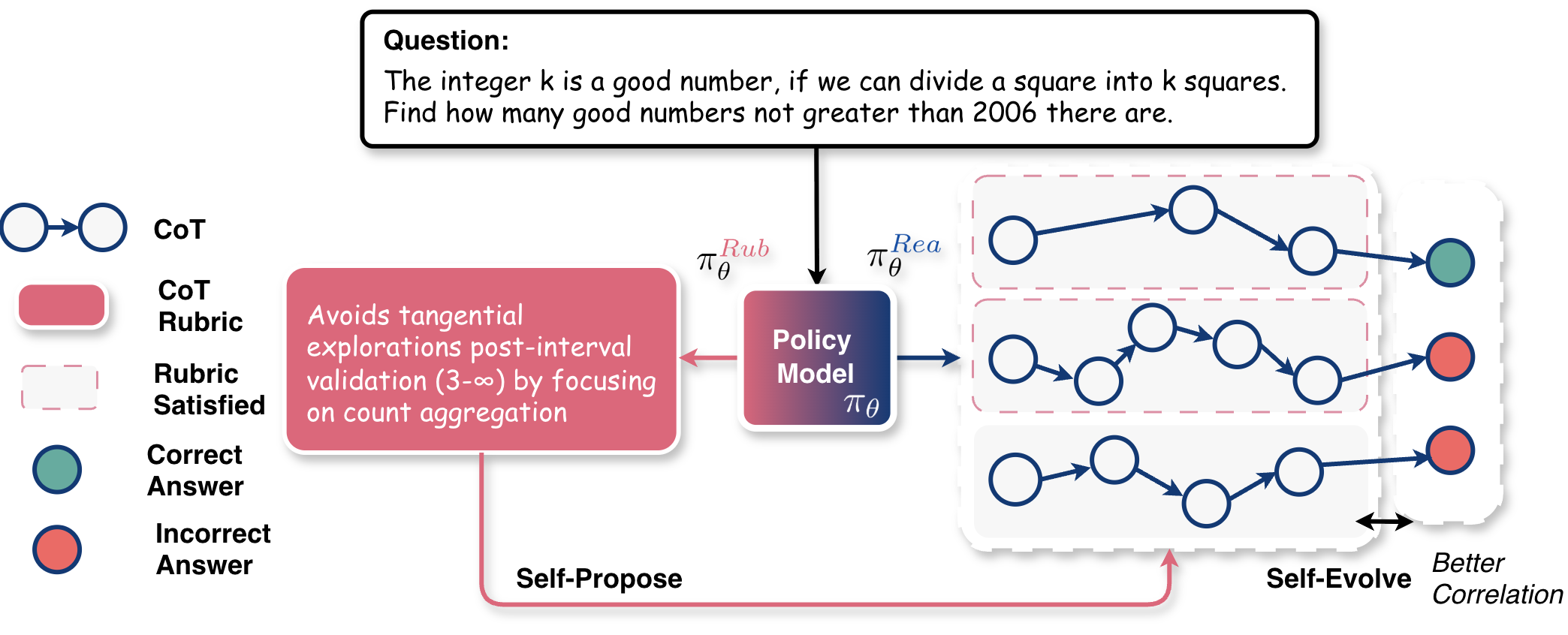}
    \vspace{-8pt}
    \caption{Key idea of reinforcement learning with CoT supervision via self-evolving rubrics (RLCER). The policy model $\pi_\theta$ acts as both the reasoner and the rubrics generator, self-generating and self-evolving the rubrics for CoT supervision, where the evolving direction is shaped towards the correlation with the rubrics satisfaction and the final answer correctness.}
    \label{fig:teaser-figure}
    \vspace{-6pt}
\end{figure}

Chain-of-thought (CoT) \cite{CoT} reasoning is essential for large language models (LLMs) to solve complex tasks, where its quality strongly affects the final-answer correctness \cite{Demystifying-CoT, what-characterizes-cot, LCoT2Tree, ProcessBench}. 
However, CoT is rarely rewarded or optimized explicitly, even often regarded as a by-product of outcome-centric optimization objectives in large-scale reinforcement learning with verifiable rewards (RLVR) \cite{Deepseek-R1, O1, DeepSeekMath}.
This yields an underconstrained learning signal: for the same final answer, many distinct reasoning trajectories receive identical rewards, allowing optimization to drift toward shortcut or brittle strategies \cite{RuscaRL, zhang2025interplay}. Consequently, the model can converge to sub-optimal reasoning behaviors, limiting robustness and overall capability \cite{PRM8K}.

\revision{However, while important, directly rewarding or supervising CoTs remains challenging in practice, mainly for two reasons:}
(i) training additional reward models (RMs) for CoT supervision typically requires labor-intensive and fine-grained annotations \cite{PRM8K, PRIME}, 
and (ii) the policy’s CoT distribution shifts during training, yielding non-stationary and potentially biased supervision. 
\textit{\textbf{Can the policy model \concept{self-propose} rubrics as CoT supervision criteria and self-evolve them during training with no human annotations?}} 
If so, this could enable a new paradigm of self-improving reasoning, shifting RL from optimizing \concept{what LLMs answer} to autonomously improving \concept{how LLMs think}, potentially yielding \concept{``free-lunch''}  reasoning gains.

We find possibilities of achieving this from the recent advances in the research line of \concept{self-evolving training methods} \cite{self-evolving-survey, AlphaZero}, where a policy model $\pi_\theta$ can improve itself with little to no human intervention during training.
On the one hand, $\pi_\theta$ itself can self-generate training data, such as verifiable question-answer pairs for RL or high-quality CoT trajectories for cold-starting reasoning \cite{AbsoluteZero, SPICE, SSP, STaR}.
On the other hand, $\pi_\theta$ itself can self-assign reward signals by acting as a judge to evaluate its own solutions \cite{Self-Rewaring-Rubric, SRLM, SeRL, SPIN}.
A key idea shared by these methods is to let a single policy model $\pi_\theta$ play multiple roles under different prompts, serving not only as a reasoner that produces solutions but also as a data generator or a reward provider \cite{AbsoluteZero, SRLM}, with all roles optimized jointly (\eg via techniques such as multi-agent reinforcement learning \cite{AbsoluteZero}). 
Together, these capabilities make it plausible to self-propose supervision criteria (\eg rubrics) for assessing CoT quality and to self-evolve as training progresses.
However, how to adaptively propose and evolve such criteria in the self-evolving training paradigm remains largely underexplored.

To fill this gap, we propose \textbf{RLCER} (\textbf{R}einforcement \textbf{L}earning with \textbf{C}oT Supervision via Self-\textbf{E}volving \textbf{R}ubrics), empowering RLVR with self-proposed and self-evolving rubrics to supervise CoT reasoning.
The core idea is to let the policy model $\pi_\theta$ self-propose CoT supervision criteria as natural-language rubrics (\ie describing desirable reasoning properties like ``avoid tangential explorations'')
\cite{Rubrics-as-Rewards, Chasing-the-Tail}, rewarding CoTs by their rubric satisfaction, and simultaneously improving the model’s rubric-generation ability throughout training.
An overview of the key process is illustrated in Figure~\ref{fig:teaser-figure}.
Specifically, a single policy model $\pi_\theta$ is instantiated with different prompts to play two roles: a reasoner $\pi_{\theta}^{\text{\textcolor{reasoner}{$Rea$}}}$ that generates CoTs $\hat{\mathcal{C}}$ and final answers $\hat{\mathcal{A}}$, and a rubricator $\pi_{\theta}^{\text{\textcolor{rubricator}{$Rub$}}}$ that generates a set of candidate rubrics $\hat{\mathcal R} \triangleq \{\hat{\tau}_k\}_{k=1}^{K}$, where $K$ is the number of rubrics $\hat{\tau}_k$. 
The reasoner is rewarded by both the final answer correctness and the CoT quality, which is measured by satisfaction of all valid rubrics $\hat{\tau}^{valid}_k$. 
Here, we deem a rubric $\hat{\tau}_k$ valid if its satisfaction indicator $\mathbf v_k$ is strongly correlated with final-answer correctness $\mathbf z$ across multiple sampled rollouts for the same question $\mathcal Q$ (\ie $\mathtt{corr}(\mathbf v_k,\mathbf z)>\alpha$), since a high correlation means that satisfying the rubric aligns with producing correct answers. 
Additionally, we encourage the rubricator to self-evolve its rubric generation capability by rewarding the fraction of valid rubrics among its proposals
 (\ie $\frac{|\{\hat{\tau}^{valid}_k\}|}{K}$). 

Generally, RLCER enables RLVR to move beyond outcome-centric optimization by rewarding CoTs with supervision criteria that are self-proposed and self-evolved, without any human annotation.
We further conduct comprehensive experiments to examine the effectiveness of the proposed RLCER. 
First, we conduct a primary study showcasing that self-proposed rubrics can be reliable and learnable even when training is driven solely by rubric-based rewards, without any outcome reward (\cf Section \ref{sec:rubrics-reliability}). 
Based on this insight, RLCER generally outperforms the naive RLVR (\cf Section \ref{sec:main-performance}). 
Additionally, introducing self-evolving training for the rubricator further improves the reasoning performance (\cf Section \ref{sec:rubrics-self-evolving}).
Moreover, the generated rubrics can even serve as in-prompt hints to enhance reasoning at inference time (\cf Section \ref{sec:rubrics-as-hint}).
Overall, our results suggest a new paradigm in which models autonomously generate CoT supervision criteria to strengthen RL training, pointing toward more self-improving and self-evolving reasoning systems. 

\section{Related Works}
\subsection{LLM Reasoning with RL}
Reinforcement learning has become an essential technique for incentivizing chain-of-thought (CoT) reasoning in LLMs \cite{RL4LRM, Survey-LLMs}.
A milestone paradigm is reinforcement learning with verifiable rewards (RLVR) \cite{DeepSeekMath, O1, Deepseek-R1}, where LLMs are trained on verifiable tasks such as math problems \cite{AIME}, puzzle solving \cite{ARC-AGI-2}, and code execution \cite{LiveCodeBench}, by rewarding final-answer correctness using task-specific verifiers.
Since such tasks are intrinsically difficult to reward-hack, RLVR tends to encourage LLMs to produce longer CoTs before answering, thereby improving reasoning performance \cite{Deepseek-R1, O1}.
However, despite its effectiveness in incentivizing reasoning, most RLVR training remains outcome-centric and provides no direct supervision over the CoT itself, lacking guidance on how the model should reason.
While several efforts have tried to mitigate this by assigning process rewards using additional reward models (RM) \cite{PRM8K}, training such RMs requires labor-intensive training data collection \cite{PRM8K, PRIME, ProcessBench}, and a static RM often fails to keep pace with the evolving CoT distribution during training \cite{Deepseek-R1}.
Therefore, such challenges hinder the combination of RLVR training with CoT supervision criteria for rewarding in practice.

\subsection{Self-Evolving in LLMs}
A critical step toward general artificial intelligence lies in the ability to self-evolve with little or no human intervention \cite{self-evolving-survey}. 
Recent work has made progress via self-training and self-play, where a single policy model $\pi_\theta$ can self-generate training data \cite{AbsoluteZero, SSP, SPICE}, or self-provide reward signals during learning \cite{SPIN, SeRL, SRLM}. 
The self-evolving capabilities can further be strengthened with techniques such as multi-role reinforcement learning, where a single policy model $\pi_\theta$ is prompted to play different roles (\eg reasoner, data generator, or reward provider) and trained with role-specific rewards \cite{AbsoluteZero, SPICE, Alignmen-Waltz}.
Therefore, these advances in such self-evolving methods suggest the possibilities of self-proposing criteria for CoT supervision, while this direction remains largely under-explored.

\subsection{Reinforcement Learning with Rubrics}
Recently, reinforcement learning with rubrics has emerged as a promising way to provide fine-grained and interpretable reward signals \cite{Rubicon, Rubrics-as-Rewards, HealthBench, Chasing-the-Tail}.
Here, rubrics are structured, checkable evaluation criteria (\ie explicit requirements for assessing outputs) \cite{Rubrics-as-Rewards} that decompose high-level objectives into multi-facet supervision signals \cite{DR-Tulu}, enabling RL to extend towards non-verifiable and open-ended domains \cite{RuscaRL, Rubrics-as-Rewards}.
However, existing work largely applies rubrics to score response quality, leaving their effectiveness for CoT supervision underexplored.
Moreover, most approaches rely on predefined or static rubrics, which may become mismatched as model behaviors evolve during training \cite{DR-Tulu}.
To address these gaps, we introduce a framework that self-proposes rubrics as CoT supervision criteria and continually evolves them throughout training, integrating rubric-based RL for CoT supervision with self-evolving training.

\section{Preliminaries}
In this section, we briefly review the necessary background that forms the foundation of our work, including incentivizing LLM reasoning with RL in Section \ref{sec:pre-training} and multi-role RL under a single policy in Section \ref{sec:pre-masrl}. 

    \subsection{Incentivizing LLM Reasoning with RL} \label{sec:pre-training}
In this work, we focus on incentivizing LLMs to reason with RL, where the policy model is optimized by rewarding the high-quality rollouts.

\textbf{Rollout.}
Given a question $\mathcal{Q}$, the policy model $\pi_{\theta}$ attempts to solve it by generating a response that contains a CoT $\hat{\mathcal{C}}$ followed by an answer $\hat{\mathcal{A}}$, which can be formulated as:
\begin{equation}
\hat{\mathcal{C}}, \hat{\mathcal{A}} \sim \pi_{\theta}(\cdot \mid\mathcal{Q}).
\end{equation}
\textbf{Rewarding.}
After rollouts generation, rewards are assigned based on response quality, which can depend on: (i) the CoT $\hat{\mathcal{C}}$ and (ii) the answer $\hat{\mathcal{A}}$.
In the widely adopted RLVR paradigm, rewards typically depend solely on the correctness of the predicted answer $\hat{\mathcal{A}}$:
\begin{equation}\label{eq:outcome_reward}
r = \psi( \mathbb{I}(\mathcal{A},\hat{\mathcal{A}})),
\end{equation}
where $\mathbb{I}(\cdot,\cdot)$ is an answer correctness indication function which returns $1$ if two answers $\mathcal{A}$ and $\hat{\mathcal{A}}$ are equivalent and $0$ otherwise, and a reward assignment function $\psi$ maps this correctness signal $\mathbb{I}(\mathcal{A},\hat{\mathcal{A}})$ to a binary reward $r$.
Through such RLVR training, the policy model $\pi_\theta$ gradually learns to answer questions, which is often associated with producing longer CoTs \cite{Deepseek-R1}.
However, under the RLVR paradigm, the CoT $\hat{\mathcal{C}}$ itself is rarely rewarded directly due to practical implementation challenges \cite{Deepseek-R1}, while the lack of such supervision may induce sub-optimal reasoning strategies \cite{PRM8K}.

\subsection{Multi-Role RL Under a Single Policy Model} \label{sec:pre-masrl}
Recently, studies have shown that a single policy model $\pi_\theta$ can effectively learn multiple roles by assigning role-specific rewards \cite{AbsoluteZero, SPIRAL}:
\begin{equation}\label{eq:reward-role}
r^{\text{role}_i} \rightarrow \pi^{\text{role}_i}_\theta,
\end{equation}
where $r^{\text{role}_i}$ denotes the reward assigned to $\text{role}_i$, and $\pi^{\text{role}_i}_\theta$ represents the same underlying policy model $\pi_\theta$ instantiated under the prompt $\mathcal{P}^{\text{role}_i}$ corresponding to that role.

A typical application under this framework is to let the model simultaneously act as a reasoner and a data generator, achieving RL training starting from no human-labeled data \cite{AbsoluteZero, AlphaZero, SSP}. 
More broadly, this multi-role, single-policy setup provides a natural substrate for self-evolving training: different roles instantiated from the same model can compete \cite{AbsoluteZero, SPICE} or collaborate \cite{Alignmen-Waltz, SPIRAL}, continuously improving the learning signals and behaviors \cite{MAS-Survey}.
Building on this view, we investigate whether the same policy can simultaneously serve as a reasoner and a CoT rubrics generator (\ie rubricator). 
With role-specific rewards, the rubricator is encouraged to produce useful CoT supervision criteria, which in turn enables the self-evolution of CoT rewarding. 
In this case, the reasoner and rubricator collaborate together for better reasoning.

\section{Methodology}

In this section, we introduce \textbf{RLCER} (\textbf{R}einforcement \textbf{L}earning with \textbf{C}oT Supervision via Self-\textbf{E}volving \textbf{R}ubrics) for bootstrapping chain-of-thought reasoning. 
In the remainder of this section, we first briefly introduce the key idea of RLCER in Section \ref{sec:key-idea} to facilitate a high-level understanding. 
Then, we discuss the multiple roles under the same policy model in Section \ref{sec:roles}. 
After that, we discuss how to incentivize reasoning with self-proposed and rubrics in Section \ref{sec:reasoner_reward}. 
We then discuss how to self-evolve such rubrics in Section \ref{sec:reward_rubricator}.
Finally, we present how to optimize the policy $\pi_\theta$ under the two roles based on role-specific rewards in Section \ref{sec:policy-optimization}.

\subsection{Key Idea} \label{sec:key-idea}
\begin{wrapfigure}{r}{0.48\textwidth}
    \centering
    \includegraphics[width=\linewidth]{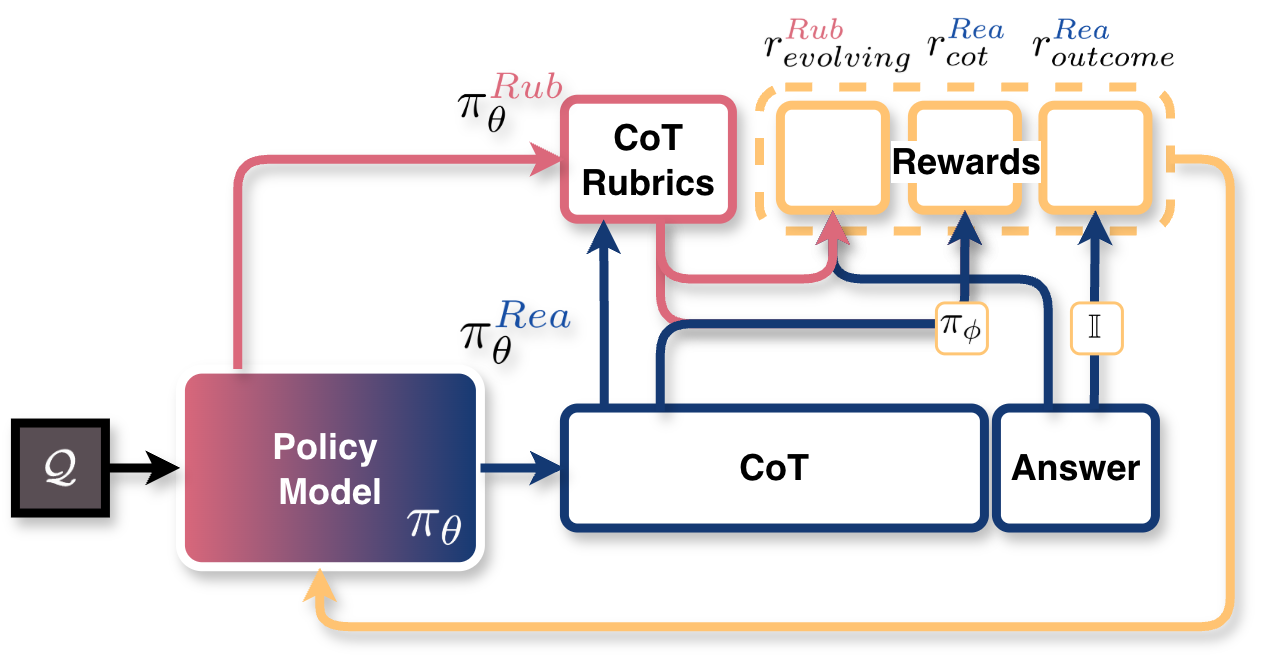}
    \vspace{-14pt}
    \caption{The RLCER loop. Format reward is ignored for brevity. One single policy model self-proposes rubrics for rewarding CoTs, and self-evolves the rubrics via rewarding generation capabilities.}
    \vspace{-8pt}
    \label{fig:key-idea}
\end{wrapfigure}

The key idea of RLCER is to go beyond rewarding what LLMs answer (\ie final-answer correctness) to explicitly and autonomously reward how LLMs think (\ie CoT quality) using self-proposed, self-evolving rubrics. The rubrics are evolved by tracking how well their satisfaction correlates with the final answer correctness.

To achieve this goal, we introduce two key roles in the multi-role RL framework for optimization: the reasoner $\pi_{\theta}^{\text{\textcolor{reasoner}{$Rea$}}}$ for answering the question and the rubricator $\pi_{\theta}^{\text{\textcolor{rubricator}{$Rub$}}}$ for generating rubrics supervising CoTs.
Here, the reasoner $\pi_{\theta}^{\text{\textcolor{reasoner}{$Rea$}}}(\cdot)=\pi_{\theta}(\cdot\mid\mathcal{P}^{\text{\textcolor{reasoner}{$Rea$}}})$ and the rubricator $\pi_{\theta}^{\text{\textcolor{rubricator}{$Rub$}}}(\cdot)=\pi_{\theta}(\cdot\mid\mathcal{P}^{\text{\textcolor{rubricator}{$Rub$}}})$ share the same policy parameter $\theta$, while their roles are distinguished by different prompts $\mathcal{P}^{\text{\textcolor{reasoner}{$Rea$}}}$ and $\mathcal{P}^{\text{\textcolor{rubricator}{$Rub$}}}$ respectively. 
The prompts can be found in Appendix \ref{apdx:prompts}. 
Additionally, a verifier $\pi_\phi$ is used for judging the satisfaction of rubrics, which is another fine-tuned and frozen model parameterized by $\phi$. 

As shown in Figure \ref{fig:key-idea}, we explicitly refine how the reasoner $\pi_{\theta}^{\text{\textcolor{reasoner}{$Rea$}}}$ thinks, by rewarding the CoT quality beyond the answer correctness, with the CoT quality evaluated by the verifier $\pi_\phi$ based on rubrics generated by the rubricator. 
We further enable rubric generation to self-evolve for more autonomous and effective CoT supervision by rewarding the quality of generated rubrics, guided by the correlation between rubric satisfaction degree and answer correctness.
We jointly optimize the policy model $\pi_\theta$ under two roles by assigning different rewards for each role. 
After computing role-specific rewards, we optimize the shared policy model $\pi_\theta$ for both roles, with advantages computed separately for each role \cite{SPICE}.

\subsection{Two Roles in One Policy: Reasoner and Rubricator} \label{sec:roles}
In this section, we take a closer look at the different roles of the reasoner $\pi_{\theta}^{\text{\textcolor{reasoner}{$Rea$}}}$ and the rubricator $\pi_{\theta}^{\text{\textcolor{rubricator}{$Rub$}}}$, within the RLCER framework.

\textbf{Reasoner} $\pi_{\theta}^{\text{\textcolor{reasoner}{$Rea$}}}$. 
The reasoner answers the question by generating a CoT $\hat{\mathcal C}$ and a final answer $\hat{\mathcal A}$.
Given the question $\mathcal Q$, the rollout sampling process of the reasoner can be formulated as:
\begin{equation}
\hat{\mathcal C}, \hat{\mathcal A} \sim 
\pi_{\theta}^{\text{\textcolor{reasoner}{$Rea$}}}\!\big(\cdot \mid \mathcal Q\big),
\end{equation}
where $\hat{\mathcal C}$ denotes the CoT process, and $\hat{\mathcal A}$ denotes the predicted final answer.

\textbf{Rubricator} $\pi_{\theta}^{\text{\textcolor{rubricator}{$Rub$}}}$.
The rubricator generates a set of $K$ rubrics for evaluating the CoT qualities based on the question $\mathcal Q$ and one CoT $\hat{\mathcal C}$. 
Given the question $\mathcal Q$ and a generated CoT $\hat{\mathcal C}$, the rollout sampling process of the rubricator can be formulated as:
\begin{equation}
\hat{\mathcal R} \sim 
\pi_{\theta}^{\text{\textcolor{rubricator}{$Rub$}}}
\big(\cdot \mid \mathcal Q, \hat{\mathcal C}\big),
\qquad
\hat{\mathcal R} \triangleq \{\hat{\tau}_k\}_{k=1}^{K},
\ \hat{\tau}_k \triangleq (\hat c_k,\hat s_k).
\end{equation}
The generated textual output $\hat{\mathcal R}$ contains $K$ specific rubrics, each denoted as $\hat{\tau}_k$. Here, $\hat c_k$ is the textual criterion (\eg ``avoids tangential explorations post-interval validation (3-$\infty$) by focusing on count aggregation'') of rubric $\hat{\tau}_k$ and $\hat s_k \in \mathbb{R}$ is the score of rubric $\hat{\tau}_k$ reflecting its importance, which can be both positive or negative. 
The rubrics generated under the rubricator role will then be used for judging the CoT quality through an external verifier $\pi_\phi$.

\subsection{Rewarding How to Think via Self-Proposed Rubrics} \label{sec:reasoner_reward}
In this section, we discuss how to autonomously reward how LLMs think via self-proposed rubrics. 

\textbf{Rewarding CoT with self-proposed rubrics.} Specifically, we reward a CoT based on how well it satisfies the valid rubrics proposed by the rubricator $\pi_{\theta}^{\text{\textcolor{rubricator}{$Rub$}}}$.
The CoT reward is computed by aggregating the scores of the satisfied rubrics:
\begin{equation}
    r_{cot}^{\text{\textcolor{reasoner}{$Rea$}}}=\texttt{norm}\left(\sum_{\hat{\tau}_{k} \in \hat{\mathcal{R}}_{valid}}\pi_\phi \left(\hat c_k, \hat{\mathcal{C}}\right)\cdot\hat s_k\right),
\end{equation}

where $\hat{\mathcal{R}}_{valid}$ denotes the set of valid rubrics from the rubricator, $\pi_\phi(\hat c_k, \hat{\mathcal{C}})\in\{0,1\}$ is a binary value produced by the verifier $\pi_\phi$ indicating whether the CoT $\hat{\mathcal{C}}$ satisfies rubric $\hat c_k$, and $\texttt{norm}(\cdot)$ is a normalization function that maps the aggregated score to the interval $[0,1]$. The function $\texttt{norm}(\cdot)$ conducts the min–max normalization as $\texttt{norm}(x) = (x-\text{MinValue}) / (\text{MaxValue}-\text{MinValue})$, where MinValue and MaxValue denote the minimum and maximum aggregated rubric scores attainable under the valid rubrics $\hat{\mathcal{R}}_{valid}$, respectively.

\textbf{Valid rubrics}. Here, a rubric $\hat{\tau}_k$ is considered valid if it is informative as a rewarding criterion: 
(i) its satisfaction indicator $\mathbf v_k$ is positively correlated with final-answer correctness $\mathbf z$
across $N$ sampled rollouts of the same question $\mathcal Q$ (\ie $\mathtt{corr}(\mathbf v_k,\mathbf z)>\alpha$),
and (ii) it is discriminative across different CoTs and would not yield consistent satisfaction (\ie $\mathtt{std}(\mathbf v_k)>0$). 
We define $\mathbf v_k$ and $\mathbf z$ over these $N$ rollouts as:
\begin{equation}
\begin{aligned}
\mathbf v_k &\triangleq \Big[\pi_\phi(\hat c_k,\hat{\mathcal C}_0),\ldots,\pi_\phi(\hat c_k,\hat{\mathcal C}_N)\Big],\\
\mathbf z   &\triangleq \Big[\mathbb{I}(\mathcal A,\hat{\mathcal A}_0),\ldots,\mathbb{I}(\mathcal A,\hat{\mathcal A}_N)\Big].
\end{aligned}
\end{equation}

The intuition is that higher correlation implies that satisfying the rubric is consistently associated with correct final answers and can produce informative rewarding signals. Here $\alpha$ is set as 0.2 by default in our paper. 

\textbf{Outcome reward.} The CoT reward $r_{cot}^{\text{\textcolor{reasoner}{$Rea$}}}$ can be used as an auxiliary reward alongside the outcome reward $r^{\text{\textcolor{reasoner}{$Rea$}}}_{outcome}$, which is calculated as follows:
\begin{equation}\label{eq:outcome_reward}
r^{\text{\textcolor{reasoner}{$Rea$}}}_{outcome} =  
\begin{cases}
  1, & \texttt{is\_equiv}(\mathcal{A},\hat{\mathcal{A}})\\
  -1, & \text{Otherwise.}
\end{cases}
\end{equation}

Following DAPO \cite{DAPO}, we assign a binary outcome reward based on final answer correctness: $r^{\text{\textcolor{reasoner}{$Rea$}}}_{outcome}$=1 if the predicted answer $\hat{\mathcal{A}}$ is equivalent to the ground truth answer $\mathcal{A}$, and $r^{\text{\textcolor{reasoner}{$Rea$}}}_{outcome}$=-1 otherwise. 

\textbf{Total reward for reasoner.} Finally, the overall reward for the reasoner is the combination of the outcome reward and the CoT reward:

\begin{equation}
r^{\text{\textcolor{reasoner}{$Rea$}}} = 
    r^{\text{\textcolor{reasoner}{$Rea$}}}_{outcome} + r_{cot}^{\text{\textcolor{reasoner}{$Rea$}}}.
\end{equation}

Such a reward enables autonomously optimizing what LLMs answer and how LLMs think simultaneously. 

\subsection{Rubrics Self-Evolving for Better Supervision} \label{sec:reward_rubricator}
While the two-role design enables a single policy model to self-propose rubrics for CoT supervision, we argue that such rubrics may get saturated: once consistently satisfied by all rollouts, they lose discriminative power over CoT quality and no longer provide reward signals, as shown in Section \ref{sec:reasoner_reward}. 
Therefore, we further propose to self-evolve the rubric generation capability, so as to progressively produce more informative and challenging rubrics that mitigate saturation.

\textbf{Self-evolving with validity reward.} To achieve rubricator self-evolving, we reward the fraction of the number of valid rubrics among all generated rubrics as follows: 
\begin{equation}
r_{evolving}^{\text{\textcolor{rubricator}{$Rub$}}}=\frac{K_{\text{valid}}}{K},
\end{equation}

where $K$ and $K_{\text{valid}} \triangleq |\{\hat{\tau}^{valid}_k\}|$ are the numbers of all generated rubrics and valid rubrics, respectively. 
Such rewards encourage the rubricator to generate more informative rubrics that are positively correlated with answer correctness (\ie $\mathtt{corr}(\mathbf v_k,\mathbf z)>\alpha$) and discriminative (\ie $\mathtt{std}(\mathbf v_k) > 0$), thereby preventing rubric saturation. 
This reward provides a simple way for evolving the quality of generated rubrics, by connecting the satisfaction of the rubrics with the final answer correctness.

\textbf{Format reward.} Additionally, to ensure the proposed rubrics can be stably parsed, we reward the rubricator with a format reward, rewarding the correct format as follows:

\begin{equation}
    r_{format}^{\text{\textcolor{rubricator}{$Rub$}}}=\begin{cases}
  1, & \text{The format is correct.}\\
  0, & \text{Otherwise.}
\end{cases}
\end{equation}

\textbf{Total reward for rubricator.} Finally, the total reward for the rubricator is calculated as the combination of the quality reward and the format reward:

\begin{equation}
   r^{\text{\textcolor{rubricator}{$Rub$}}} =  r_{evolving}^{\text{\textcolor{rubricator}{$Rub$}}} + r_{format}^{\text{\textcolor{rubricator}{$Rub$}}}.
\end{equation}

Such a reward design encourages the rubricator to continuously refine its proposed rubrics, consistently providing informative rewarding signals for CoT supervision.  

\begin{figure*}[t]
    \captionsetup{justification=raggedright,singlelinecheck=false}
    \centering
    \scalebox{1}[1]{%
        \includegraphics[width=0.88\textwidth]{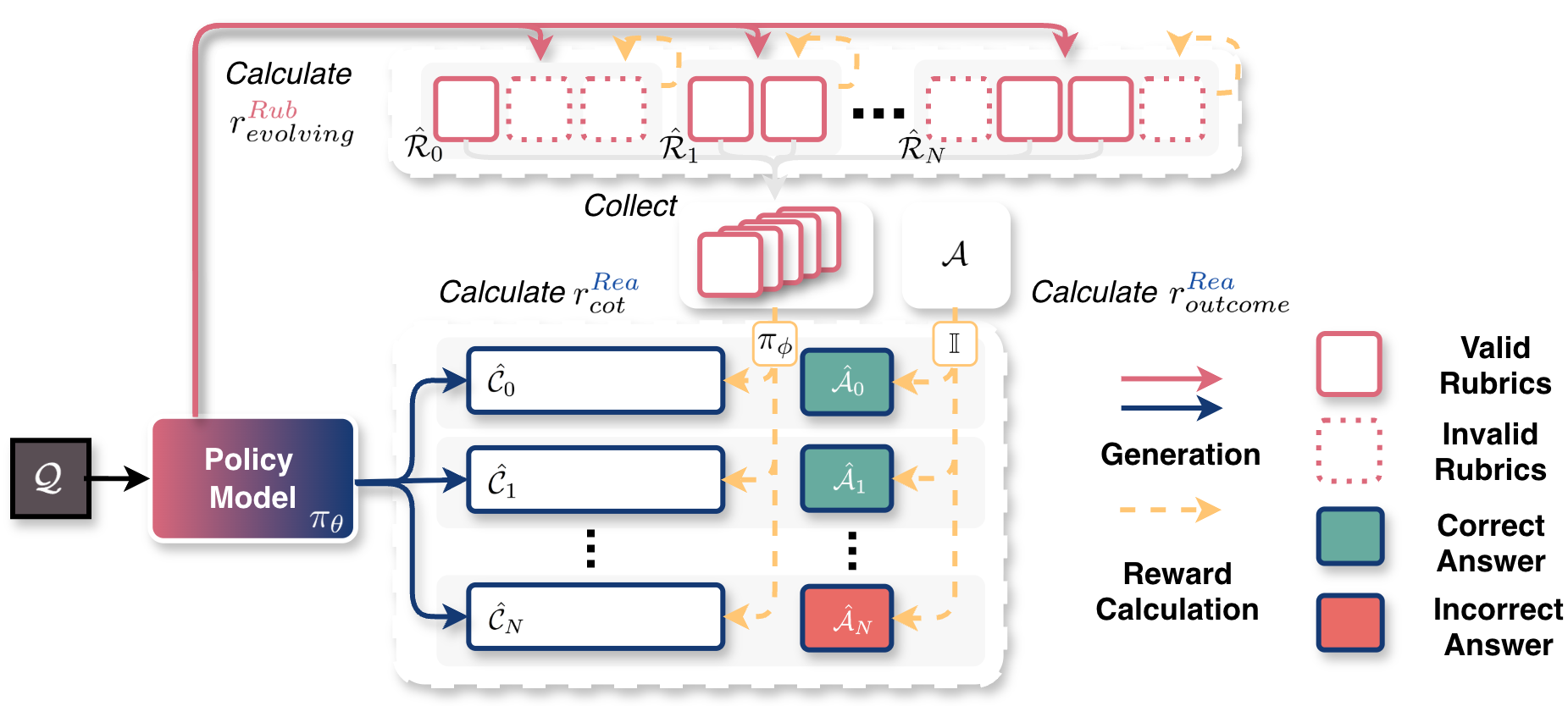}%
    }
    \vspace{-6pt}
    \caption{Illustration of the reward calculation process in RLCER. For question $\mathcal{Q}$, the reasoner generates $N$ responses each with CoT $\hat{\mathcal{C}}_k$ and the final answer $\hat{\mathcal{A}}_k$. 
    After that, the rubricator generates $K_{n}$ specific rubrics (\ie $\hat{\mathcal R}_n \triangleq \{\hat{\tau}_k\}_{k=1}^{K_n}$). 
    The outcome reward is applied first by matching the generated answer $\hat{\mathcal{A}}_k$ with the ground-truth answer $\mathcal{A}$.
    All the valid rubrics (\ie $\mathtt{corr}(\mathbf v_k,\mathbf z)>\alpha$ and $\mathtt{std}(\mathbf v_k)>0$) are collected for rewarding CoTs. 
    And the fraction of valid rubrics for the $k$-th rubricator generation (\ie $\frac{|\{\hat{\tau}^{valid}_{n,k}\}|}{K_n}$) is used for rewarding the rubricator to self-evolve. 
    }
    \label{fig:main-figure}
\end{figure*}

\subsection{Two-Role Optimization under a Single Policy} \label{sec:policy-optimization}

After computing the rewards for the reasoner and the rubricator, we jointly optimize the shared policy model $\pi_\theta$ under both roles in an end-to-end manner. Specifically, we compute role-specific advantages $\hat{A}^{\text{\textcolor{reasoner}{$Rea$}}}_t$ and $\hat{A}^{\text{\textcolor{rubricator}{$Rub$}}}_t$  from their respective rewards $r^{\text{\textcolor{reasoner}{$Rea$}}}$ and $r^{\text{\textcolor{rubricator}{$Rub$}}}$, and use them to update the policy via a standard policy-gradient objective. Gradients from both roles are aggregated to update the same set of parameters $\theta$, allowing the model to learn from role-specific feedback while maintaining a unified policy. The overall optimization procedure is illustrated in Algorithm~\ref{def:aglo}.

With role-specific advantages $\hat{A}^{\text{\textcolor{reasoner}{$Rea$}}}_t$ and $\hat{A}^{\text{\textcolor{rubricator}{$Rub$}}}_t$, the optimization objective can be formulated as follows: 
\begin{equation}
\begin{aligned}
\mathcal{J}(\theta)
&= \ 
\mathbb{E}_{(\mathcal{Q}, \mathcal{A}) \sim \mathcal{D}^{\text{\textcolor{reasoner}{$Rea$}}},
\ \boldsymbol{o} \sim \pi^{\text{\textcolor{reasoner}{$Rea$}}}_{\theta_{\text{old}}}(\cdot \mid \mathcal{Q})}
\Big[
\min\Big(
\rho_t(\theta)\,\hat{A}^{\text{\textcolor{reasoner}{$Rea$}}}_t,\;
\text{clip}\big(
\rho_t(\theta),\,
1-\epsilon,\,
1+\epsilon
\big)\,\hat{A}^{\text{\textcolor{reasoner}{$Rea$}}}_t
\Big)
\Big],
\\
&+\mathbb{E}_{(\mathcal{Q}, \hat{\mathcal C}) \sim \mathcal{D}^{\text{\textcolor{rubricator}{$Rub$}}},
\ \boldsymbol{o} \sim \pi^{\text{\textcolor{rubricator}{$Rub$}}}_{\theta_{\text{old}}}(\cdot \mid \mathcal{Q, \hat{\mathcal C}})}
\Big[
\min\Big(
\rho_t(\theta)\,\hat{A}^{\text{\textcolor{rubricator}{$Rub$}}}_t,\;
\text{clip}\big(
\rho_t(\theta),\,
1-\epsilon,\,
1+\epsilon
\big)\,\hat{A}^{\text{\textcolor{rubricator}{$Rub$}}}_t
\Big)
\Big],
\end{aligned}
\end{equation}

where $\rho_t(\theta)$ is the policy likelihood ratio between $\pi_\theta$ and $\pi_{\theta_{\text{old}}}$ at step $t$, and $\epsilon$ is the clipping hyperparameter (\eg $0.2$). 
It is worth noting that GRPO is unsuitable for our algorithm, since the rubricator operates under different contexts, preventing its rollouts from being grouped under the same context required by GRPO \cite{DeepSeekMath}.

\begin{algorithm}[t]
\captionsetup{labelformat=default,labelsep=colon,name=Algorithm}
   \caption{Policy Optimization with RLCER}
   \label{def:aglo}
\begin{algorithmic}
   \STATE {\bfseries Input:} Question $\mathcal{Q}$, answer $\mathcal{A}$, the rollout sampling number $N$, and the maximum training steps $T$.
   \STATE {\bfseries Initialization:} initialize policy model $\pi_\theta$, training step $t \leftarrow 0$.
  \WHILE{$t \leq T$}
    \STATE \tikzmk{rlcerRolloutA} \CommentBelow{sample reasoner rollouts}
    \STATE $\{\hat{\mathcal{C}}_n, \hat{\mathcal{A}}_n\}_{n=1}^N \sim  \pi_{\theta}^{\text{\textcolor{reasoner}{$Rea$}}}(\cdot\mid\mathcal{Q})$
    \STATE \CommentBelow{sample rubricator rollouts}
    \STATE $\{\hat{\mathcal{R}}_n\}_{n=1}^N \sim  \pi_{\theta}^{\text{\textcolor{rubricator}{$Rub$}}}(\cdot\mid\mathcal{Q}, \hat{\mathcal{C}_n})$\tikzmk{rlcerRolloutB}
    \STATE \tikzmk{rlcerRewardA} \CommentBelow{calculate reasoner rewards}
    \STATE $\{r_n^{\text{\textcolor{reasoner}{$Rea$}}}\}_{n=1}^N =
    \{r^{\text{\textcolor{reasoner}{$Rea$}}}_{outcome,n} + r_{cot,n}^{\text{\textcolor{reasoner}{$Rea$}}}\}_{n=1}^N $
    \STATE \CommentBelow{calculate rubricator rewards}
    \STATE $\{r_n^{\text{\textcolor{rubricator}{$Rub$}}}\}_{n=1}^N =  \{r_{evolving,n}^{\text{\textcolor{rubricator}{$Rub$}}} + r_{format,n}^{\text{\textcolor{rubricator}{$Rub$}}}\}_{n=1}^N $\tikzmk{rlcerRewardB}
    \STATE \tikzmk{rlcerPolicyA} \CommentBelow{calculate role-specifc advantages}
    \STATE Calculate $\hat{A}^{\text{\textcolor{reasoner}{$Rea$}}}_t$ and $\hat{A}^{\text{\textcolor{rubricator}{$Rub$}}}_t$
    \STATE \CommentBelow{policy optimization}
    \STATE Update policy $\pi_\theta$ via $\mathcal{J}(\theta)$\tikzmk{rlcerPolicyB}

    \STATE $t \leftarrow t+1$
   \ENDWHILE

   \STATE {\bfseries return} $\pi_\theta$.
\end{algorithmic}

\boxit[Role-specific rollout]{RolloutBg}{rlcerRolloutA}{rlcerRolloutB}
\boxit[Rewarding]{RewardBg}{rlcerRewardA}{rlcerRewardB}
\boxit[Policy updating]{PolicyBg}{rlcerPolicyA}{rlcerPolicyB}
\end{algorithm}
\section{Experiments}

\begin{itemize}[leftmargin=*]
    \item \textbf{RQ1:} Can self-proposed rubrics provide reliable rewarding signals for RL training?
    \item \textbf{RQ2:} How does RLCER enhance the reasoning capabilities of LRMs?
    \item \textbf{RQ3:} How does the self-evolving enable the rubricator to gradually propose better rubrics?
    \item \textbf{RQ4:} How effective are generated rubrics at facilitating LLM reasoning when used as in-prompt hints?
\end{itemize}

\subsection{Experiment Setup}
We briefly introduce the experiment setup of our paper, and more details can be found in Appendix \ref{apdx:impl-details}.

\textbf{LLMs.} 
We conduct experiments on two open-source Qwen models of different sizes (8B and 4B) \cite{Qwen3}.
We first cold-start the pre-trained models (\ie Qwen3-8B-Base and Qwen3-4B-Base) by performing supervised fine-tuning on correctly formatted data obtained via reject sampling from a stronger teacher model, as we observe that even post-trained versions of such small-sized models (\eg Qwen3-8B) cannot always strictly follow the required instructions when acting as the rubricator and reasoner. 
More details can be found in Appendix \ref{apdx:impl-details}.
All subsequent RL experiments are conducted based on these carefully cold-started models.

\textbf{Training and evaluation.} 
We train models based on the DAPO-Math-17k dataset \cite{DAPO}, which contains 17k high-quality math questions. 
We evaluate models on various reasoning benchmarks.
For math reasoning, we evaluate math-reasoning benchmarks including AIME24 \cite{AIME}, AIME25 \cite{AIME}, and AMC23 \cite{aimo_amc_2023}.
For general knowledge reasoning, we evaluate on GPQA-Diamond \cite{GPQA} and SuperGPQA \cite{SuperGPQA}. 
For the large size of SuperGPQA, we evaluate on three subsets, each including 100 questions, namely SuperGPQA-Eng, SuperGPQA-Med, and SuperGPQA-Sci. 
For each testing question, we sample $N$=16 responses with sampling temperature as 0.7.
We report pass@1 as the accuracy metric (\ie the average pass rate among 16 responses).
For all RL training, we fix the max training steps to 1500.

\subsection{(RQ1) Reliability of Self-Proposed Rubrics} \label{sec:rubrics-reliability}

\begin{wrapfigure}{r}{0.52\textwidth}
  \vspace{-10pt}
  \centering
  \begin{subfigure}[ht]{0.49\linewidth}
    \centering
    \includegraphics[width=\linewidth]{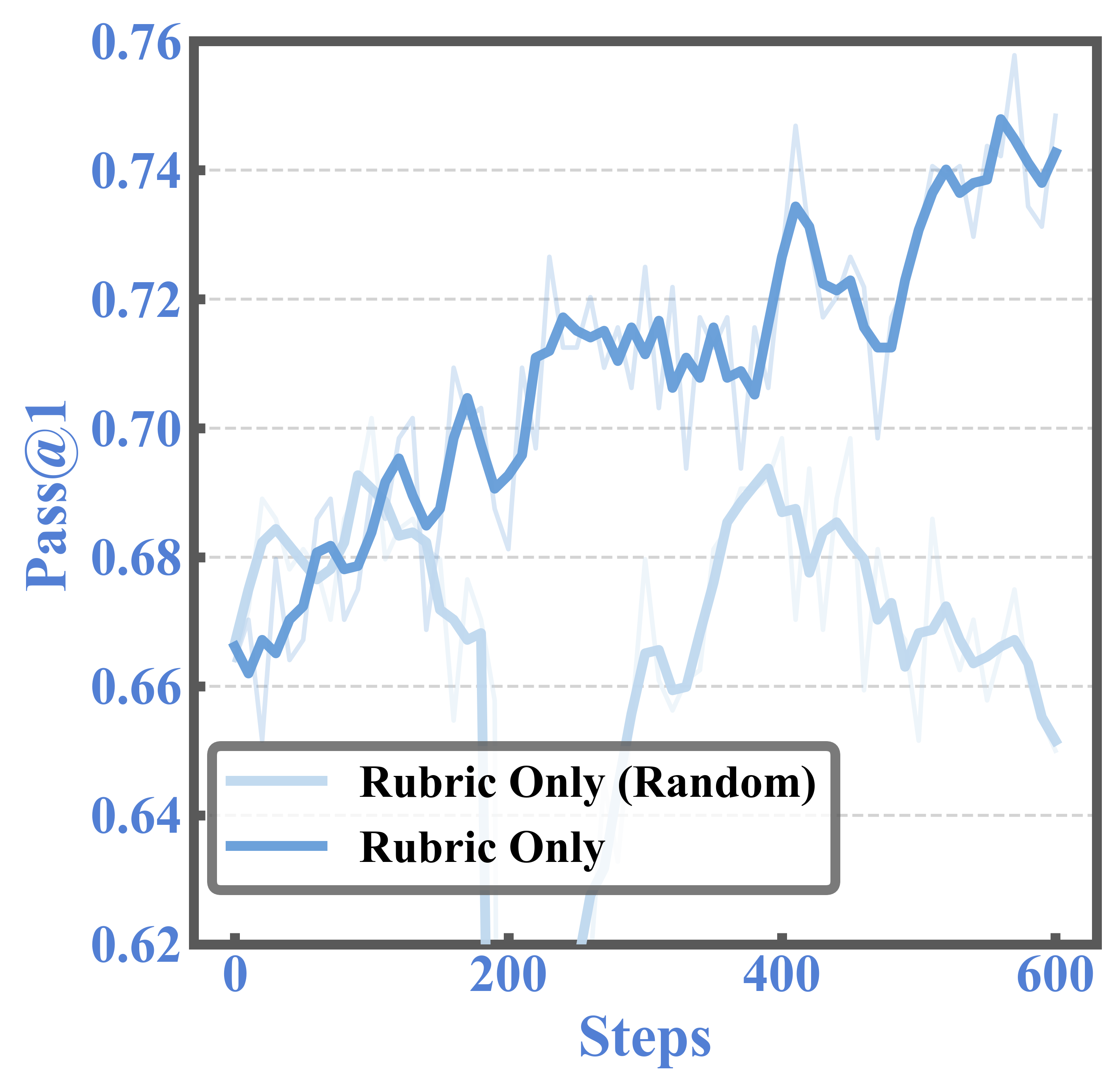}
    \caption{Accuracy on AMC23.}
    \label{fig:rubric-only-a}
  \end{subfigure}\hfill
  \begin{subfigure}[ht]{0.49\linewidth}
    \centering
    \includegraphics[width=\linewidth]{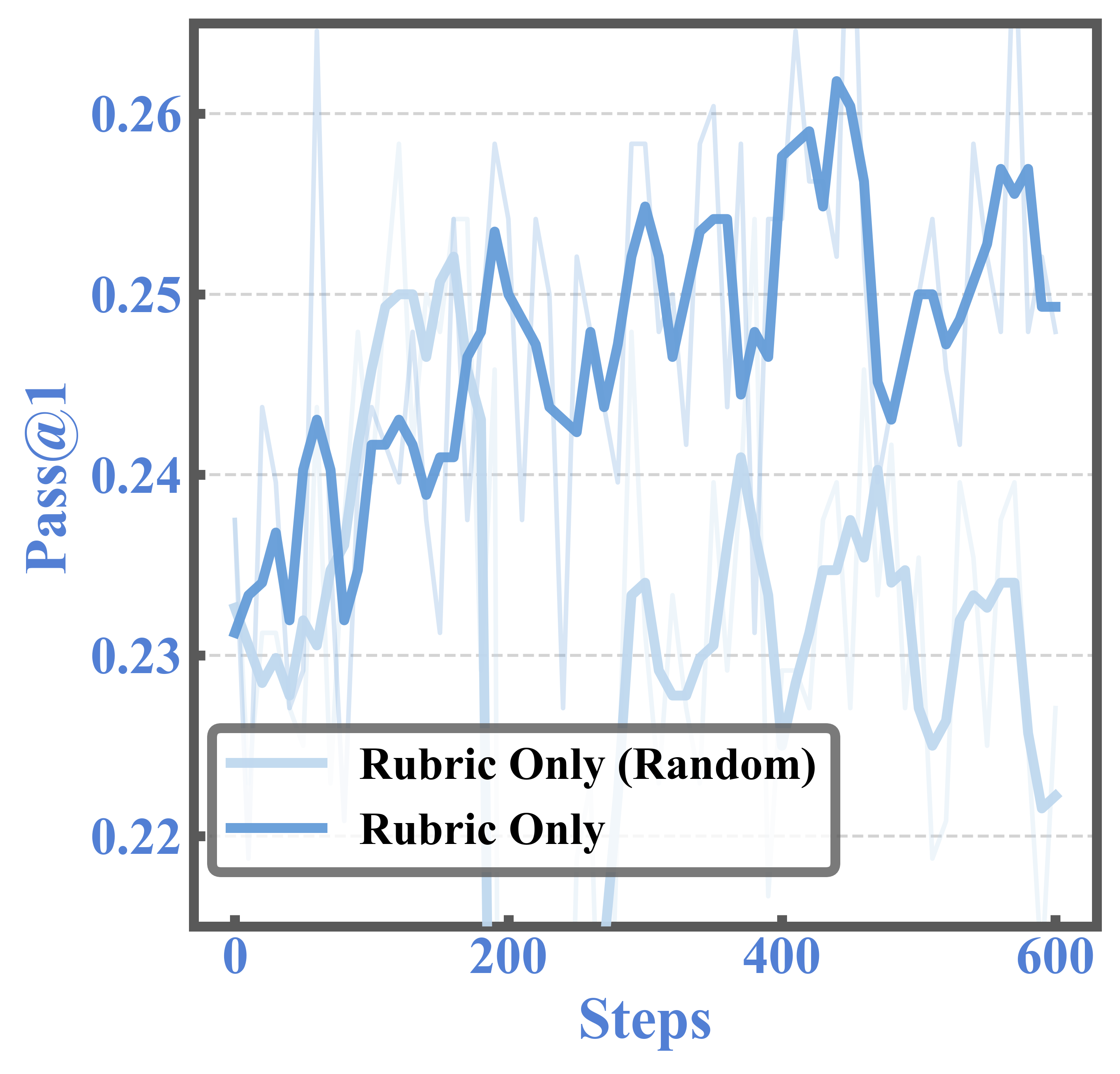}
    \caption{Accuracy on AIME25.}
    \label{fig:rubric-only-b}
  \end{subfigure}
  \vspace{-10pt}
  \caption{Accuracy dynamics when only rewarding CoTs. Even when rewarding CoTs with self-proposed rubrics without any outcome reward can bring consistent performance gain.}
  \label{fig:rubric-only}
  \vspace{-10pt}
\end{wrapfigure}

A key factor that determines the success of our method lies in whether the self-proposed and self-evolving rubrics can provide meaningful learning signals for RL. 
To verify this, we first conduct a preliminary experiment, considering an outcome-free setting that trains the model by \textbf{only rewarding CoTs based on self-proposed rubrics without any outcome rewards} (\ie $r^{\text{\textcolor{reasoner}{$Rea$}}} =  r_{cot}^{\text{\textcolor{reasoner}{$Rea$}}}$). 
If the model still learns to reason effectively under this setup, it would indicate that self-proposed rubrics can yield meaningful RL signals. 
We report the accuracy dynamics on reasoning tasks during the training in Figure \ref{fig:rubric-only}, where the bold lines are the rolling averaged performance with a window size of 3 and the slim and shallow lines are the original performance. 
We have the following observation:

\textbf{Self-proposed rubrics can provide meaningful CoT rewarding signals for RL training. }
As shown in this figure, by only rewarding CoTs with the self-proposed rubrics (\ie Rubric Only in the figure), the reasoning capability of the LLM shows consistent improvement during training. 
In contrast, when we assign random values between 0 and 1 as rubric rewards (\ie Rubric Only (Random) in the figure), the LLM fails to achieve performance improvements and even exhibits sudden performance drops during training around 200 steps. 
Such results show that the self-proposed and self-evolving rubrics can yield meaningful CoT rewarding signals, which lay an important foundation for our approach RLCER.

\subsection{(RQ2) Effectiveness of RLCER} \label{sec:main-performance}
Based on the reliability of self-proposed rubrics, we further test the effectiveness of combining both $r^{\text{\textcolor{reasoner}{$Rea$}}}_{outcome}$ and $r_{cot}^{\text{\textcolor{reasoner}{$Rea$}}}$ in RLCER. 
We compare RLCER with the vanilla RLVR which only considers the outcome reward, and report the performance comparison in Table \ref{tab:main-table}. 
Here, both RLCER and RLVR start from the SFT checkpoint.
We have the following observations:

\textbf{RLCER outperforms the outcome-centric RLVR training. }
As shown in Table \ref{tab:main-table}, training with RLCER outperforms the vanilla RLVR across multiple datasets and multiple model sizes on average. 
Even when trained only on math datasets, RLCER also generalizes well on general reasoning tasks (\ie GPQA series.)
Additionally, we observe that the performance gain is more obvious on larger size models (\ie 8B models benefit more from RLCER than 4B models). 
Such phenomena validate the effectiveness of RLCER, which provides a ``free-lunch'' for improving the RL performance, by enhancing it with self-proposed and self-evolving rubrics. 
Case studies about the generated rubrics can be found in Appendix \ref{apdx:case-study}.

\newcommand{\EMainTableScale}{0.99 }

\begin{table*}[!t]
\centering
\caption{The performance comparison across diverse reasoning benchmarks.}

\makebox[\linewidth][c]{%
  \resizebox{0.92\linewidth}{!}{%
    \begin{tabular}{cl|ccccccc}
    \toprule
    \multicolumn{2}{c|}{\textbf{}} &
    \multicolumn{7}{c}{\textbf{Datasets}} \\
    \textbf{Scale} & \textbf{Method} &
    \textbf{AIME2024} & \textbf{AIME2025} & \textbf{AMC2023} &
    \textbf{GPQA-Diamond} &
    \textbf{SuperGPQA-Eng} & \textbf{SuperGPQA-Med} & \textbf{SuperGPQA-Sci} \\
    \midrule

    \multirow{4}{*}{4B}
    & Base
    & 11.25 & 6.46 & 31.09 & 7.77 & 18.50 & 14.25 & 15.44 \\
    & SFT
    & 17.29 & 18.96 & 59.53 & 24.43 & 31.69 & 28.25 & 27.81 \\
    \cmidrule(lr){2-9}
    & + RLVR
    & 29.38 & 30.21 & 79.53 & 44.16 & 39.75 & 28.50 & \textbf{42.88} \\
    & \cellcolor{lightgray}+ RLCER
    & \cellcolor{lightgray}\textbf{29.38}
    & \cellcolor{lightgray}\textbf{30.63}
    & \cellcolor{lightgray}\textbf{81.88}
    & \cellcolor{lightgray}\textbf{44.91}
    & \cellcolor{lightgray}\textbf{40.19}
    & \cellcolor{lightgray}\textbf{31.63}
    & \cellcolor{lightgray}41.81 \\
    \midrule\midrule

    \multirow{4}{*}{8B}
    & Base
    & 11.46 & 9.58 & 42.34 & 24.37 & 33.06 & 24.69 & 23.63 \\
    & SFT
    & 22.29 & 23.75 & 66.41 & 31.72 & 36.00 & 33.75 & 35.19 \\
    \cmidrule(lr){2-9}
    & + RLVR
    & 34.79 & 32.50 & 84.53 & 46.56 & 42.94 & \textbf{38.31} & 48.81 \\
    & \cellcolor{lightgray}+ RLCER
    & \cellcolor{lightgray}\textbf{37.50}
    & \cellcolor{lightgray}\textbf{33.33}
    & \cellcolor{lightgray}\textbf{86.41}
    & \cellcolor{lightgray}\textbf{48.77}
    & \cellcolor{lightgray}\textbf{45.00}
    & \cellcolor{lightgray}36.50
    & \cellcolor{lightgray}\textbf{50.25} \\

    \bottomrule
    \end{tabular}
  }%
}

\label{tab:main-table}
\end{table*}

\subsection{(RQ3) Mechanism of Rubrics Self-Evolving} \label{sec:rubrics-self-evolving}

In this section, we study how the rubrics self-evolving affects training within the RLCER framework (\ie the role of the reward $r_{evolving}^{\text{\textcolor{rubricator}{$Rub$}}}$). 
To study this, we conduct an ablation study on the 8B-sized model by abandoning the evolving reward $r_{evolving}^{\text{\textcolor{rubricator}{$Rub$}}}$, and the rubricator is only rewarded with the format reward. 
We report the training dynamics in Figure \ref{fig:ablation-perf} and Figure \ref{fig:ablation-study-wrap} including the averaged performance (\ie averaged accuracy among AIME2024, AIME2025, and AMC2023), the averaged correlation between rubrics satisfaction and final answer correctness (\ie $\mathtt{corr}(\mathbf v_k,\mathbf z)$), and the CoT reward of the reasoner (\ie $r_{cot}^{\text{\textcolor{reasoner}{$Rea$}}}$). 
The performance is reported as the rolling average of 5 scores for a clearer demonstration, where the bold lines report the rolling average scores and the shallow and slim lines report the raw scores.
We have the following observations:

\begin{wrapfigure}{r}{0.52\textwidth}
  \vspace{-10pt}
  \centering

  \begin{subfigure}[t]{0.48\linewidth}
    \centering
    \includegraphics[width=\linewidth]{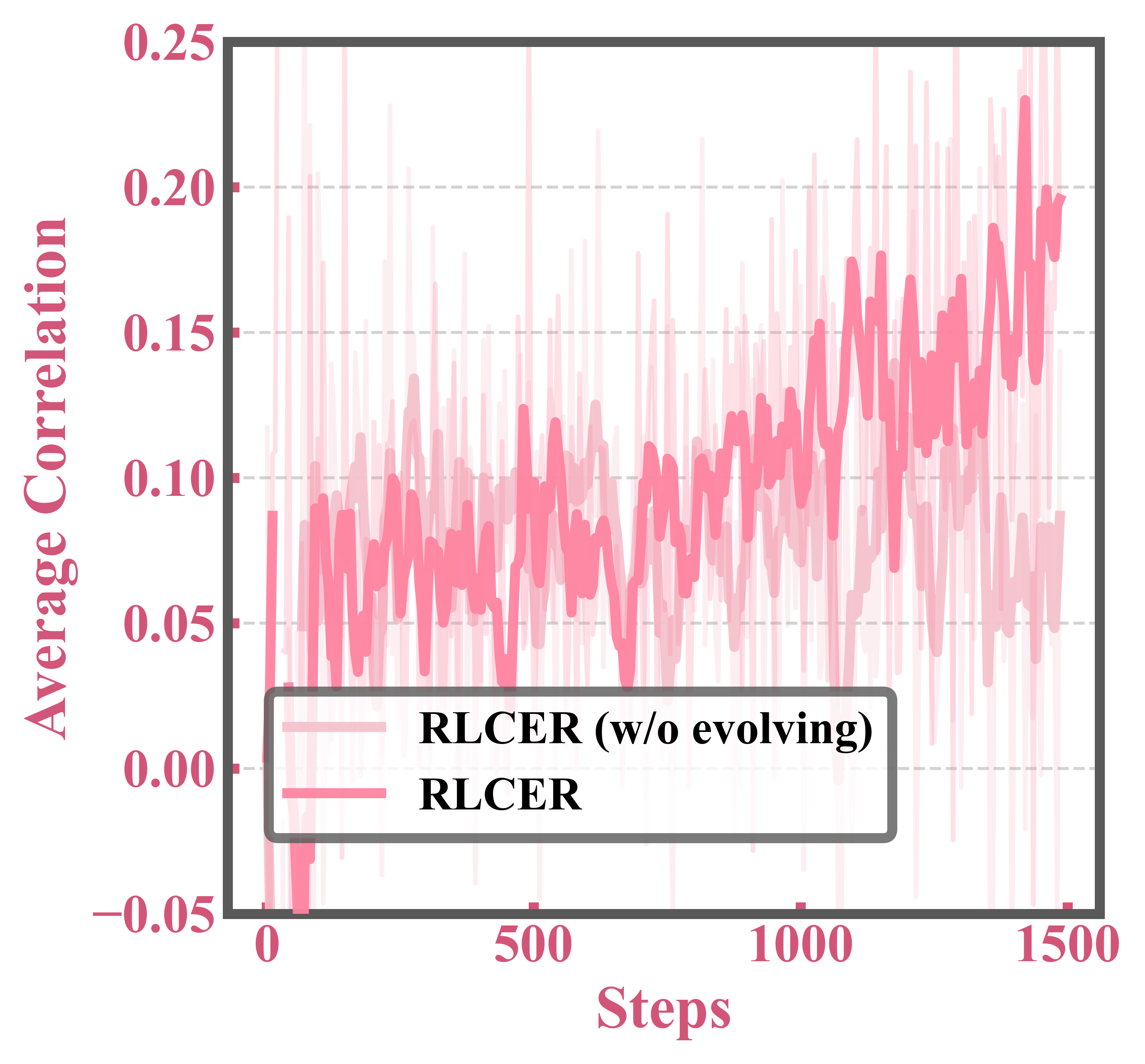}
    \subcaption{Correlation dynamics.}
    \label{fig:ablation-a}
  \end{subfigure}\hfill
  \begin{subfigure}[t]{0.463\linewidth}
    \centering
    \includegraphics[width=\linewidth]{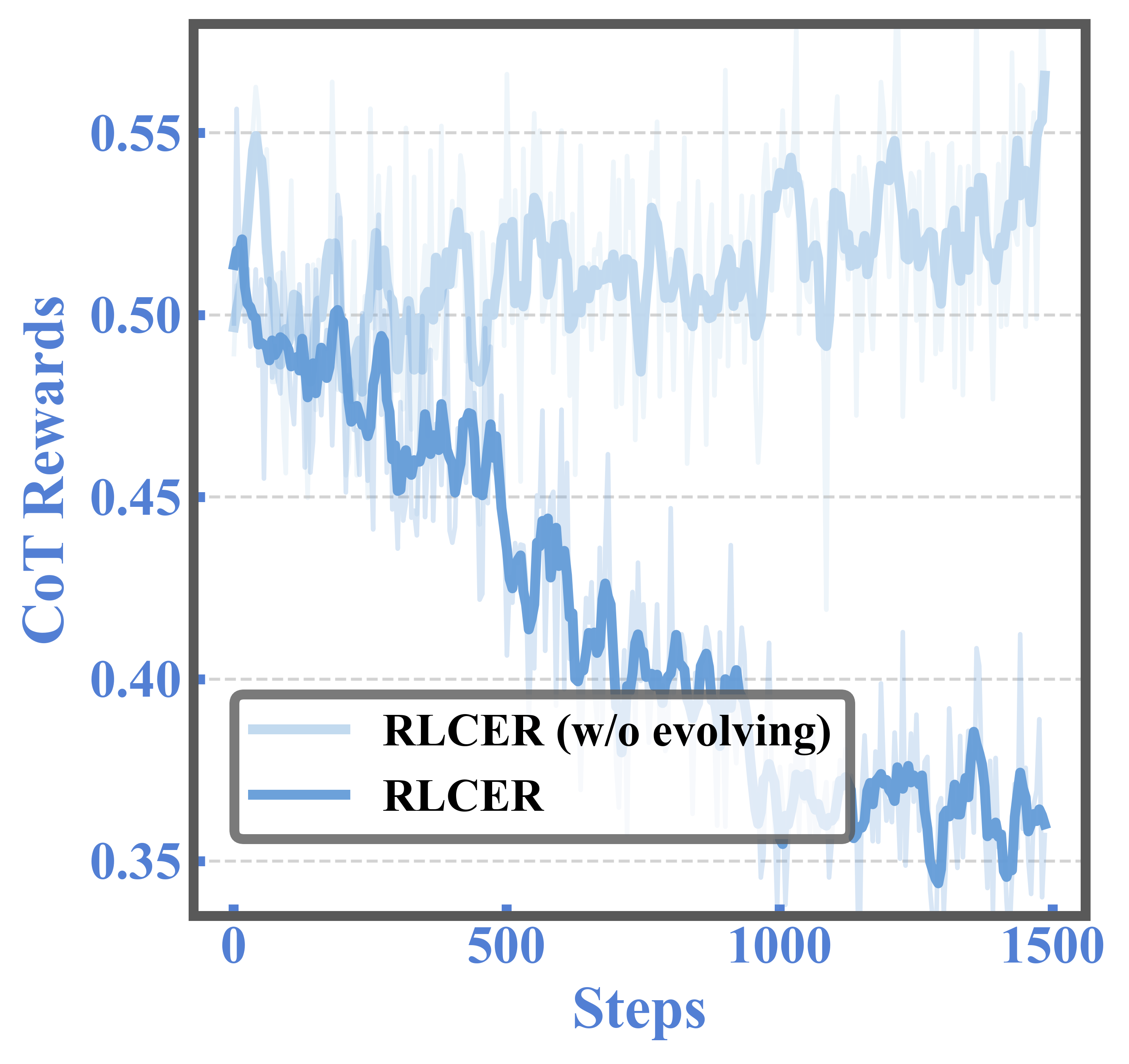}
    \subcaption{$r_{cot}^{\text{\textcolor{reasoner}{$Rea$}}}$ dynamics.}
    \label{fig:ablation-b}
  \end{subfigure}

  \vspace{-8pt}
  \caption{The effect of self-evolving on correlation and CoT-reward dynamics.}
  \label{fig:ablation-study-wrap}
  \vspace{-16pt}
\end{wrapfigure}

\textbf{Self-evolving rubrics enable better performance. }
As shown in Figure \ref{fig:ablation-perf}, by rewarding with $r_{evolving}^{\text{\textcolor{rubricator}{$Rub$}}}$, RLCER exhibits a more stable learning curve and gradually outperforms the ablation version (\ie RLCER (w/o evolving)). 
Additionally, both RLCER and RLCER (w/o evolving) outperform the naive RLVR method as it quickly converges to a sub-optimal performance.
Such phenomena show the effectiveness of self-evolving in enhancing reasoning. 

\textbf{Self-evolving with $r_{evolving}^{\text{\textcolor{rubricator}{$Rub$}}}$ enables more informative rubrics. }
We further analyze whether self-evolving enables the policy model to propose better rubrics.
As shown in Figure \ref{fig:ablation-a}, the averaged correlation between rubric satisfaction and final answer correctness (\ie $\mathtt{corr}(\mathbf v_k,\mathbf z)$) increases along the training process, while the ablation baseline shows an unchanged correlation. 
This indicates that self-evolving enables the policy model to gradually propose more informative rubrics that better align with the final answer correctness. 
Additionally, with such evolved rubrics, it gradually becomes harder to gain the CoT reward $r_{evolving}^{\text{\textcolor{rubricator}{$Rub$}}}$ as the Figure \ref{fig:ablation-b} shows a decreasing trend during the RLCER training. 
While the ablation baseline’s rubrics can be gradually satisfied by $\pi_{\theta}^{\text{\textcolor{reasoner}{$Rea$}}}$ as $r_{evolving}^{\text{\textcolor{rubricator}{$Rub$}}}$ increases. Such results highlight the importance of rubrics self-evolving.

\subsection{(RQ4) Effectiveness of Rubrics as In-prompt Reasoning Hints} \label{sec:rubrics-as-hint}

\begin{wrapfigure}{r}{0.52\textwidth}
  \vspace{-10pt}
  \centering
  \includegraphics[width=\linewidth]{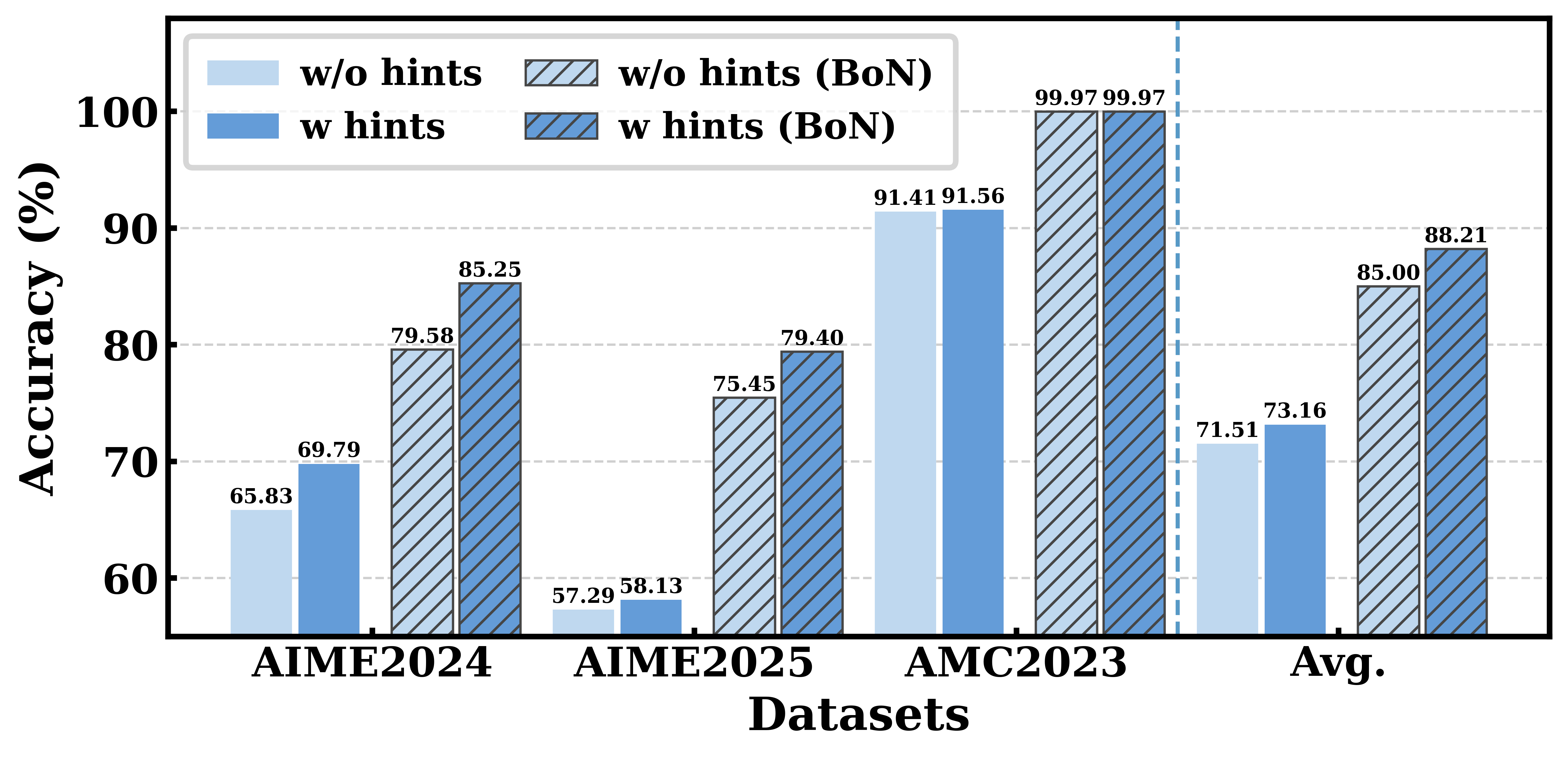}
  \vspace{-20pt}
  \caption{Performance with rubrics as in-prompt hints.}
  \label{fig:rubrics-as-hint}
  \vspace{-10pt}
\end{wrapfigure}

In this section, we attempt to understand how the generated rubrics can guide the model toward better reasoning. 
While we have verified that self-proposed and self-evolved rubrics can generate valid signals for rewarding CoTs, the mechanism through which these rubrics implicitly steer reasoning processes remains elusive, largely because the underlying reward signals are difficult to analyze.
To this end, we explicitly employ these self-proposed rubrics as in-prompt hints, and evaluate the model’s reasoning performance under this setup. 
If incorporating the rubrics into the prompt yields performance improvements, it demonstrates that the rubrics offer valid guidance for the reasoning path. 
We report the performance of Qwen3-8B with a max response length of 20480 in Figure \ref{fig:rubrics-as-hint}. 
As shown in the figure, when using generated rubrics as in-prompt hints, the performance improves, indicating that these rubrics can indeed guide better reasoning. 
Additionally, the performance of BoN (N=16) improved even further on the AIME datasets, indicating that the rubrics introduced as hints may incentivize LLMs to conduct more exploration.

\renewcommand{\arraystretch}{0.9}

\section{Limitations}
There are still several limitations in this paper.
On the one hand, the introduction of the rubricator role increases the rollout burden and thereby requires more training time.
On the other hand, our method is still quite limited to the RLVR domain, leaving the effectiveness of rewarding with self-proposed rubrics on non-verifiable domains unknown.

\section{Conclusion}
Outcome-centric RLVR improves final-answer accuracy but largely ignores direct supervision for CoTs, risking suboptimal reasoning strategies. 
However, directly rewarding CoT is challenging since reward models are costly to train and brittle under distribution shift and reward hacking. 
To address these issues, we proposed RLCER, which rewards CoTs with self-proposed, self-evolving rubrics without human annotation. 
Experiments showed that RLCER consistently outperforms outcome-only RLVR across datasets and LLM backbones, highlighting that such self-proposed and self-evolving rubrics can provide meaningful CoT supervision signals pushing the upper bound of LLM reasoning. 
Our future work will include exploring the generalization of RLCER to non-verifiable domains.

\clearpage

\bibliographystyle{plainnat}
\bibliography{main}

@inproceedings{PRM8K,
  author       = {Hunter Lightman and
                  Vineet Kosaraju and
                  Yuri Burda and
                  Harrison Edwards and
                  Bowen Baker and
                  Teddy Lee and
                  Jan Leike and
                  John Schulman and
                  Ilya Sutskever and
                  Karl Cobbe},
  title        = {Let's Verify Step by Step},
  booktitle    = {{ICLR}},
  year         = {2024}
}

@article{PRIME,
  author       = {Ganqu Cui and
                  Lifan Yuan and
                  Zefan Wang and
                  Hanbin Wang and
                  Wendi Li and
                  Bingxiang He and
                  Yuchen Fan and
                  Tianyu Yu and
                  Qixin Xu and
                  Weize Chen and
                  Jiarui Yuan and
                  Huayu Chen and
                  Kaiyan Zhang and
                  Xingtai Lv and
                  Shuo Wang and
                  Yuan Yao and
                  Xu Han and
                  Hao Peng and
                  Yu Cheng and
                  Zhiyuan Liu and
                  Maosong Sun and
                  Bowen Zhou and
                  Ning Ding},
  title        = {Process Reinforcement through Implicit Rewards},
  journal      = {CoRR},
  volume       = {abs/2502.01456},
  year         = {2025}
}

@inproceedings{DAPO,
  author       = {Qiying Yu and
                  Zheng Zhang and
                  Ruofei Zhu and
                  Yufeng Yuan and
                  Xiaochen Zuo and
                  Yu Yue and
                  Tiantian Fan and
                  Gaohong Liu and
                  Lingjun Liu and
                  Xin Liu and
                  Haibin Lin and
                  Zhiqi Lin and
                  Bole Ma and
                  Guangming Sheng and
                  Yuxuan Tong and
                  Chi Zhang and
                  Mofan Zhang and
                  Wang Zhang and
                  Hang Zhu and
                  Jinhua Zhu and
                  Jiaze Chen and
                  Jiangjie Chen and
                  Chengyi Wang and
                  Hongli Yu and
                  Weinan Dai and
                  Yuxuan Song and
                  Xiangpeng Wei and
                  Hao Zhou and
                  Jingjing Liu and
                  Wei{-}Ying Ma and
                  Ya{-}Qin Zhang and
                  Lin Yan and
                  Mu Qiao and
                  Yonghui Wu and
                  Mingxuan Wang},
  title        = {{DAPO:} An Open-Source {LLM} Reinforcement Learning System at Scale},
  booktitle    = {NeurIPS},
  year         = {2025}
}

@article{Survey-LLMs,
  author       = {Wayne Xin Zhao and
                  Kun Zhou and
                  Junyi Li and
                  Tianyi Tang and
                  Xiaolei Wang and
                  Yupeng Hou and
                  Yingqian Min and
                  Beichen Zhang and
                  Junjie Zhang and
                  Zican Dong and
                  Yifan Du and
                  Chen Yang and
                  Yushuo Chen and
                  Zhipeng Chen and
                  Jinhao Jiang and
                  Ruiyang Ren and
                  Yifan Li and
                  Xinyu Tang and
                  Zikang Liu and
                  Peiyu Liu and
                  Jian{-}Yun Nie and
                  Ji{-}Rong Wen},
  title        = {A Survey of Large Language Models},
  journal      = {CoRR},
  volume       = {abs/2303.18223},
  year         = {2023}
}

@article{Deepseek-R1,
  author       = {DeepSeek{-}AI and
                  Daya Guo and
                  Dejian Yang and
                  Haowei Zhang and
                  Junxiao Song and
                  Ruoyu Zhang and
                  Runxin Xu and
                  Qihao Zhu and
                  Shirong Ma and
                  Peiyi Wang and
                  Xiao Bi and
                  Xiaokang Zhang and
                  Xingkai Yu and
                  Yu Wu and
                  Z. F. Wu and
                  Zhibin Gou and
                  Zhihong Shao and
                  Zhuoshu Li and
                  Ziyi Gao and
                  Aixin Liu and
                  Bing Xue and
                  Bingxuan Wang and
                  Bochao Wu and
                  Bei Feng and
                  Chengda Lu and
                  Chenggang Zhao and
                  Chengqi Deng and
                  Chenyu Zhang and
                  Chong Ruan and
                  Damai Dai and
                  Deli Chen and
                  Dongjie Ji and
                  Erhang Li and
                  Fangyun Lin and
                  Fucong Dai and
                  Fuli Luo and
                  Guangbo Hao and
                  Guanting Chen and
                  Guowei Li and
                  H. Zhang and
                  Han Bao and
                  Hanwei Xu and
                  Haocheng Wang and
                  Honghui Ding and
                  Huajian Xin and
                  Huazuo Gao and
                  Hui Qu and
                  Hui Li and
                  Jianzhong Guo and
                  Jiashi Li and
                  Jiawei Wang and
                  Jingchang Chen and
                  Jingyang Yuan and
                  Junjie Qiu and
                  Junlong Li and
                  J. L. Cai and
                  Jiaqi Ni and
                  Jian Liang and
                  Jin Chen and
                  Kai Dong and
                  Kai Hu and
                  Kaige Gao and
                  Kang Guan and
                  Kexin Huang and
                  Kuai Yu and
                  Lean Wang and
                  Lecong Zhang and
                  Liang Zhao and
                  Litong Wang and
                  Liyue Zhang and
                  Lei Xu and
                  Leyi Xia and
                  Mingchuan Zhang and
                  Minghua Zhang and
                  Minghui Tang and
                  Meng Li and
                  Miaojun Wang and
                  Mingming Li and
                  Ning Tian and
                  Panpan Huang and
                  Peng Zhang and
                  Qiancheng Wang and
                  Qinyu Chen and
                  Qiushi Du and
                  Ruiqi Ge and
                  Ruisong Zhang and
                  Ruizhe Pan and
                  Runji Wang and
                  R. J. Chen and
                  R. L. Jin and
                  Ruyi Chen and
                  Shanghao Lu and
                  Shangyan Zhou and
                  Shanhuang Chen and
                  Shengfeng Ye and
                  Shiyu Wang and
                  Shuiping Yu and
                  Shunfeng Zhou and
                  Shuting Pan and
                  S. S. Li},
  title        = {DeepSeek-R1: Incentivizing Reasoning Capability in LLMs via Reinforcement
                  Learning},
  journal      = {CoRR},
  volume       = {abs/2501.12948},
  year         = {2025}
}

@article{Qwen3,
  author       = {An Yang and
                  Anfeng Li and
                  Baosong Yang and
                  Beichen Zhang and
                  Binyuan Hui and
                  Bo Zheng and
                  Bowen Yu and
                  Chang Gao and
                  Chengen Huang and
                  Chenxu Lv and
                  Chujie Zheng and
                  Dayiheng Liu and
                  Fan Zhou and
                  Fei Huang and
                  Feng Hu and
                  Hao Ge and
                  Haoran Wei and
                  Huan Lin and
                  Jialong Tang and
                  Jian Yang and
                  Jianhong Tu and
                  Jianwei Zhang and
                  Jian Yang and
                  Jiaxi Yang and
                  Jingren Zhou and
                  Junyang Lin and
                  Kai Dang and
                  Keqin Bao and
                  Kexin Yang and
                  Le Yu and
                  Lianghao Deng and
                  Mei Li and
                  Mingfeng Xue and
                  Mingze Li and
                  Pei Zhang and
                  Peng Wang and
                  Qin Zhu and
                  Rui Men and
                  Ruize Gao and
                  Shixuan Liu and
                  Shuang Luo and
                  Tianhao Li and
                  Tianyi Tang and
                  Wenbiao Yin and
                  Xingzhang Ren and
                  Xinyu Wang and
                  Xinyu Zhang and
                  Xuancheng Ren and
                  Yang Fan and
                  Yang Su and
                  Yichang Zhang and
                  Yinger Zhang and
                  Yu Wan and
                  Yuqiong Liu and
                  Zekun Wang and
                  Zeyu Cui and
                  Zhenru Zhang and
                  Zhipeng Zhou and
                  Zihan Qiu},
  title        = {Qwen3 Technical Report},
  journal      = {CoRR},
  volume       = {abs/2505.09388},
  year         = {2025}
}

@article{O1,
  author       = {Aaron Jaech and
                  Adam Kalai and
                  Adam Lerer and
                  Adam Richardson and
                  Ahmed El{-}Kishky and
                  Aiden Low and
                  Alec Helyar and
                  Aleksander Madry and
                  Alex Beutel and
                  Alex Carney and
                  Alex Iftimie and
                  Alex Karpenko and
                  Alex Tachard Passos and
                  Alexander Neitz and
                  Alexander Prokofiev and
                  Alexander Wei and
                  Allison Tam and
                  Ally Bennett and
                  Ananya Kumar and
                  Andre Saraiva and
                  Andrea Vallone and
                  Andrew Duberstein and
                  Andrew Kondrich and
                  Andrey Mishchenko and
                  Andy Applebaum and
                  Angela Jiang and
                  Ashvin Nair and
                  Barret Zoph and
                  Behrooz Ghorbani and
                  Ben Rossen and
                  Benjamin Sokolowsky and
                  Boaz Barak and
                  Bob McGrew and
                  Borys Minaiev and
                  Botao Hao and
                  Bowen Baker and
                  Brandon Houghton and
                  Brandon McKinzie and
                  Brydon Eastman and
                  Camillo Lugaresi and
                  Cary Bassin and
                  Cary Hudson and
                  Chak Ming Li and
                  Charles de Bourcy and
                  Chelsea Voss and
                  Chen Shen and
                  Chong Zhang and
                  Chris Koch and
                  Chris Orsinger and
                  Christopher Hesse and
                  Claudia Fischer and
                  Clive Chan and
                  Dan Roberts and
                  Daniel Kappler and
                  Daniel Levy and
                  Daniel Selsam and
                  David Dohan and
                  David Farhi and
                  David Mely and
                  David Robinson and
                  Dimitris Tsipras and
                  Doug Li and
                  Dragos Oprica and
                  Eben Freeman and
                  Eddie Zhang and
                  Edmund Wong and
                  Elizabeth Proehl and
                  Enoch Cheung and
                  Eric Mitchell and
                  Eric Wallace and
                  Erik Ritter and
                  Evan Mays and
                  Fan Wang and
                  Felipe Petroski Such and
                  Filippo Raso and
                  Florencia Leoni and
                  Foivos Tsimpourlas and
                  Francis Song and
                  Fred von Lohmann and
                  Freddie Sulit and
                  Geoff Salmon and
                  Giambattista Parascandolo and
                  Gildas Chabot and
                  Grace Zhao and
                  Greg Brockman and
                  Guillaume Leclerc and
                  Hadi Salman and
                  Haiming Bao and
                  Hao Sheng and
                  Hart Andrin and
                  Hessam Bagherinezhad and
                  Hongyu Ren and
                  Hunter Lightman and
                  Hyung Won Chung and
                  Ian Kivlichan and
                  Ian O'Connell and
                  Ian Osband and
                  Ignasi Clavera Gilaberte and
                  Ilge Akkaya},
  title        = {OpenAI o1 System Card},
  journal      = {CoRR},
  volume       = {abs/2412.16720},
  year         = {2024}
}

@article{RL4LRM,
  author       = {Kaiyan Zhang and
                  Yuxin Zuo and
                  Bingxiang He and
                  Youbang Sun and
                  Runze Liu and
                  Che Jiang and
                  Yuchen Fan and
                  Kai Tian and
                  Guoli Jia and
                  Pengfei Li and
                  Yu Fu and
                  Xingtai Lv and
                  Yuchen Zhang and
                  Sihang Zeng and
                  Shang Qu and
                  Haozhan Li and
                  Shijie Wang and
                  Yuru Wang and
                  Xinwei Long and
                  Fangfu Liu and
                  Xiang Xu and
                  Jiaze Ma and
                  Xuekai Zhu and
                  Ermo Hua and
                  Yihao Liu and
                  Zonglin Li and
                  Huayu Chen and
                  Xiaoye Qu and
                  Yafu Li and
                  Weize Chen and
                  Zhenzhao Yuan and
                  Junqi Gao and
                  Dong Li and
                  Zhiyuan Ma and
                  Ganqu Cui and
                  Zhiyuan Liu and
                  Biqing Qi and
                  Ning Ding and
                  Bowen Zhou},
  title        = {A Survey of Reinforcement Learning for Large Reasoning Models},
  journal      = {CoRR},
  volume       = {abs/2509.08827},
  year         = {2025}
}

@article{Self-Rewaring-Rubric,
  author       = {Zhiling Ye and
                  Yun Yue and
                  Haowen Wang and
                  Xudong Han and
                  Jiadi Jiang and
                  Cheng Wei and
                  Lei Fan and
                  Jiaxin Liang and
                  Shuowen Zhang and
                  Ji Li and
                  Chunxiao Guo and
                  Jian Wang and
                  Peng Wei and
                  Jinjie Gu},
  title        = {Self-Rewarding Rubric-Based Reinforcement Learning for Open-Ended
                  Reasoning},
  journal      = {CoRR},
  volume       = {abs/2509.25534},
  year         = {2025}
}

@inproceedings{SRLM,
  author       = {Weizhe Yuan and
                  Richard Yuanzhe Pang and
                  Kyunghyun Cho and
                  Xian Li and
                  Sainbayar Sukhbaatar and
                  Jing Xu and
                  Jason Weston},
  title        = {Self-Rewarding Language Models},
  booktitle    = {{ICML}},
  year         = {2024}
}

@inproceedings{STaR,
  author       = {Eric Zelikman and
                  Yuhuai Wu and
                  Jesse Mu and
                  Noah D. Goodman},
  title        = {STaR: Bootstrapping Reasoning With Reasoning},
  booktitle    = {NeurIPS},
  year         = {2022}
}

@article{AbsoluteZero,
  author       = {Andrew Zhao and
                  Yiran Wu and
                  Yang Yue and
                  Tong Wu and
                  Quentin Xu and
                  Yang Yue and
                  Matthieu Lin and
                  Shenzhi Wang and
                  Qingyun Wu and
                  Zilong Zheng and
                  Gao Huang},
  title        = {Absolute Zero: Reinforced Self-play Reasoning with Zero Data},
  journal      = {CoRR},
  volume       = {abs/2505.03335},
  year         = {2025}
}

@article{SPICE,
  author       = {Bo Liu and
                  Chuanyang Jin and
                  Seungone Kim and
                  Weizhe Yuan and
                  Wenting Zhao and
                  Ilia Kulikov and
                  Xian Li and
                  Sainbayar Sukhbaatar and
                  Jack Lanchantin and
                  Jason Weston},
  title        = {{SPICE:} Self-Play In Corpus Environments Improves Reasoning},
  journal      = {CoRR},
  volume       = {abs/2510.24684},
  year         = {2025}
}

@article{AlphaZero,
  author       = {David Silver and
                  Thomas Hubert and
                  Julian Schrittwieser and
                  Ioannis Antonoglou and
                  Matthew Lai and
                  Arthur Guez and
                  Marc Lanctot and
                  Laurent Sifre and
                  Dharshan Kumaran and
                  Thore Graepel and
                  Timothy P. Lillicrap and
                  Karen Simonyan and
                  Demis Hassabis},
  title        = {Mastering Chess and Shogi by Self-Play with a General Reinforcement
                  Learning Algorithm},
  journal      = {CoRR},
  volume       = {abs/1712.01815},
  year         = {2017}
}

@article{DR-Tulu,
  title={DR Tulu: Reinforcement Learning with Evolving Rubrics for Deep Research},
  author={Shao, Rulin and Asai, Akari and Shen, Shannon Zejiang and Ivison, Hamish and Kishore, Varsha and Zhuo, Jingming and Zhao, Xinran and Park, Molly and Finlayson, Samuel G and Sontag, David and others},
  journal={arXiv preprint arXiv:2511.19399},
  year={2025}
}

@article{SSP,
  author       = {Hongliang Lu and
                  Yuhang Wen and
                  Pengyu Cheng and
                  Ruijin Ding and
                  Haotian Xu and
                  Jiaqi Guo and
                  Chutian Wang and
                  Haonan Chen and
                  Xiaoxi Jiang and
                  Guanjun Jiang},
  title        = {Search Self-play: Pushing the Frontier of Agent Capability without
                  Supervision},
  journal      = {CoRR},
  volume       = {abs/2510.18821},
  year         = {2025}
}

@article{Rubicon,
  author       = {Zenan Huang and
                  Yihong Zhuang and
                  Guoshan Lu and
                  Zeyu Qin and
                  Haokai Xu and
                  Tianyu Zhao and
                  Ru Peng and
                  Jiaqi Hu and
                  Zhanming Shen and
                  Xiaomeng Hu and
                  Xijun Gu and
                  Peiyi Tu and
                  Jiaxin Liu and
                  Wenyu Chen and
                  Yuzhuo Fu and
                  Zhiting Fan and
                  Yanmei Gu and
                  Yuanyuan Wang and
                  Zhengkai Yang and
                  Jianguo Li and
                  Junbo Zhao},
  title        = {Reinforcement Learning with Rubric Anchors},
  journal      = {CoRR},
  volume       = {abs/2508.12790},
  year         = {2025}
}

@article{GPQA,
  author       = {David Rein and
                  Betty Li Hou and
                  Asa Cooper Stickland and
                  Jackson Petty and
                  Richard Yuanzhe Pang and
                  Julien Dirani and
                  Julian Michael and
                  Samuel R. Bowman},
  title        = {{GPQA:} {A} Graduate-Level Google-Proof Q{\&}A Benchmark},
  journal      = {CoRR},
  volume       = {abs/2311.12022},
  year         = {2023}
}

@misc{AIME,
  author = {{Mathematical Association of America}},
  title = {American Invitational Mathematics Examination (AIME)},
  year = {2024},
  month = {February},
  url = {https://maa.org/math-competitions/american-invitational-mathematics-examination-aime}
}

@article{HealthBench,
  author       = {Rahul K. Arora and
                  Jason Wei and
                  Rebecca Soskin Hicks and
                  Preston Bowman and
                  Joaquin Qui{\~{n}}onero Candela and
                  Foivos Tsimpourlas and
                  Michael Sharman and
                  Meghan Shah and
                  Andrea Vallone and
                  Alex Beutel and
                  Johannes Heidecke and
                  Karan Singhal},
  title        = {HealthBench: Evaluating Large Language Models Towards Improved Human
                  Health},
  journal      = {CoRR},
  volume       = {abs/2505.08775},
  year         = {2025}
}

@inproceedings{ProcessBench,
  author       = {Chujie Zheng and
                  Zhenru Zhang and
                  Beichen Zhang and
                  Runji Lin and
                  Keming Lu and
                  Bowen Yu and
                  Dayiheng Liu and
                  Jingren Zhou and
                  Junyang Lin},
  title        = {ProcessBench: Identifying Process Errors in Mathematical Reasoning},
  booktitle    = {ACL},
  year         = {2025}
}

@article{what-characterizes-cot,
  author       = {Yunzhen Feng and
                  Julia Kempe and
                  Cheng Zhang and
                  Parag Jain and
                  Anthony Hartshorn},
  title        = {What Characterizes Effective Reasoning? Revisiting Length, Review,
                  and Structure of CoT},
  journal      = {CoRR},
  volume       = {abs/2509.19284},
  year         = {2025}
}

@article{LCoT2Tree,
  author       = {Gangwei Jiang and
                  Yahui Liu and
                  Zhaoyi Li and
                  Qi Wang and
                  Fuzheng Zhang and
                  Linqi Song and
                  Ying Wei and
                  Defu Lian},
  title        = {What Makes a Good Reasoning Chain? Uncovering Structural Patterns
                  in Long Chain-of-Thought Reasoning},
  journal      = {CoRR},
  volume       = {abs/2505.22148},
  year         = {2025}
}

@article{Demystifying-CoT,
  author       = {Edward Y. Chang and
                  Yuxuan Tong and
                  Morry Niu and
                  Graham Neubig and
                  Xiang Yue},
  title        = {Demystifying Long Chain-of-Thought Reasoning in LLMs},
  journal      = {CoRR},
  volume       = {abs/2502.03373},
  year         = {2025}
}

@article{Rubrics-as-Rewards,
  author       = {Anisha Gunjal and
                  Anthony Wang and
                  Elaine Lau and
                  Vaskar Nath and
                  Bing Liu and
                  Sean Hendryx},
  title        = {Rubrics as Rewards: Reinforcement Learning Beyond Verifiable Domains},
  journal      = {CoRR},
  volume       = {abs/2507.17746},
  year         = {2025}
}

@article{RuscaRL,
  author       = {Yang Zhou and
                  Sunzhu Li and
                  Shunyu Liu and
                  Wenkai Fang and
                  Jiale Zhao and
                  Jingwen Yang and
                  Jianwei Lv and
                  Kongcheng Zhang and
                  Yihe Zhou and
                  Hengtong Lu and
                  Wei Chen and
                  Yan Xie and
                  Mingli Song},
  title        = {Breaking the Exploration Bottleneck: Rubric-Scaffolded Reinforcement
                  Learning for General {LLM} Reasoning},
  journal      = {CoRR},
  volume       = {abs/2508.16949},
  year         = {2025}
}

@article{Chasing-the-Tail,
  author       = {Junkai Zhang and
                  Zihao Wang and
                  Lin Gui and
                  Swarnashree Mysore Sathyendra and
                  Jaehwan Jeong and
                  Victor Veitch and
                  Wei Wang and
                  Yunzhong He and
                  Bing Liu and
                  Lifeng Jin},
  title        = {Chasing the Tail: Effective Rubric-based Reward Modeling for Large
                  Language Model Post-Training},
  journal      = {CoRR},
  volume       = {abs/2509.21500},
  year         = {2025}
}

@inproceedings{CoT,
  author       = {Jason Wei and
                  Xuezhi Wang and
                  Dale Schuurmans and
                  Maarten Bosma and
                  Brian Ichter and
                  Fei Xia and
                  Ed H. Chi and
                  Quoc V. Le and
                  Denny Zhou},
  title        = {Chain-of-Thought Prompting Elicits Reasoning in Large Language Models},
  booktitle    = {NeurIPS},
  year         = {2022}
}

@article{DeepSeekMath,
  author       = {Zhihong Shao and
                  Peiyi Wang and
                  Qihao Zhu and
                  Runxin Xu and
                  Junxiao Song and
                  Mingchuan Zhang and
                  Y. K. Li and
                  Y. Wu and
                  Daya Guo},
  title        = {DeepSeekMath: Pushing the Limits of Mathematical Reasoning in Open
                  Language Models},
  journal      = {CoRR},
  volume       = {abs/2402.03300},
  year         = {2024}
}

@article{self-evolving-survey,
  author       = {Huan{-}ang Gao and
                  Jiayi Geng and
                  Wenyue Hua and
                  Mengkang Hu and
                  Xinzhe Juan and
                  Hongzhang Liu and
                  Shilong Liu and
                  Jiahao Qiu and
                  Xuan Qi and
                  Yiran Wu and
                  Hongru Wang and
                  Han Xiao and
                  Yuhang Zhou and
                  Shaokun Zhang and
                  Jiayi Zhang and
                  Jinyu Xiang and
                  Yixiong Fang and
                  Qiwen Zhao and
                  Dongrui Liu and
                  Qihan Ren and
                  Cheng Qian and
                  Zhenhailong Wang and
                  Minda Hu and
                  Huazheng Wang and
                  Qingyun Wu and
                  Heng Ji and
                  Mengdi Wang},
  title        = {A Survey of Self-Evolving Agents: On Path to Artificial Super Intelligence},
  journal      = {CoRR},
  volume       = {abs/2507.21046},
  year         = {2025}
}

@article{SeRL,
  author       = {Wenkai Fang and
                  Shunyu Liu and
                  Yang Zhou and
                  Kongcheng Zhang and
                  Tongya Zheng and
                  Kai{-}Xuan Chen and
                  Mingli Song and
                  Dacheng Tao},
  title        = {SeRL: Self-Play Reinforcement Learning for Large Language Models with
                  Limited Data},
  journal      = {CoRR},
  volume       = {abs/2505.20347},
  year         = {2025}
}

@article{ARC-AGI-2,
  author       = {Fran{\c{c}}ois Chollet and
                  Mike Knoop and
                  Gregory Kamradt and
                  Bryan Landers and
                  Henry Pinkard},
  title        = {{ARC-AGI-2:} {A} New Challenge for Frontier {AI} Reasoning Systems},
  journal      = {CoRR},
  volume       = {abs/2505.11831},
  year         = {2025}
}

@inproceedings{LiveCodeBench,
  author       = {Naman Jain and
                  King Han and
                  Alex Gu and
                  Wen{-}Ding Li and
                  Fanjia Yan and
                  Tianjun Zhang and
                  Sida Wang and
                  Armando Solar{-}Lezama and
                  Koushik Sen and
                  Ion Stoica},
  title        = {LiveCodeBench: Holistic and Contamination Free Evaluation of Large
                  Language Models for Code},
  booktitle    = {{ICLR}},
  publisher    = {OpenReview.net},
  year         = {2025}
}

@inproceedings{SPIN,
  author       = {Zixiang Chen and
                  Yihe Deng and
                  Huizhuo Yuan and
                  Kaixuan Ji and
                  Quanquan Gu},
  title        = {Self-Play Fine-Tuning Converts Weak Language Models to Strong Language
                  Models},
  booktitle    = {{ICML}},
  publisher    = {OpenReview.net},
  year         = {2024}
}

@article{Alignmen-Waltz,
  author       = {Jingyu Zhang and
                  Haozhu Wang and
                  Eric Michael Smith and
                  Sid Wang and
                  Amr Sharaf and
                  Mahesh Pasupuleti and
                  Benjamin Van Durme and
                  Daniel Khashabi and
                  Jason Weston and
                  Hongyuan Zhan},
  title        = {The Alignment Waltz: Jointly Training Agents to Collaborate for Safety},
  journal      = {CoRR},
  volume       = {abs/2510.08240},
  year         = {2025}
}

@article{MAS-Survey,
  author       = {Khanh{-}Tung Tran and
                  Dung Dao and
                  Minh{-}Duong Nguyen and
                  Quoc{-}Viet Pham and
                  Barry O'Sullivan and
                  Hoang D. Nguyen},
  title        = {Multi-Agent Collaboration Mechanisms: {A} Survey of LLMs},
  journal      = {CoRR},
  volume       = {abs/2501.06322},
  year         = {2025}
}

@article{SPIRAL,
  author       = {Mickel Liu and
                  Liwei Jiang and
                  Yancheng Liang and
                  Simon Shaolei Du and
                  Yejin Choi and
                  Tim Althoff and
                  Natasha Jaques},
  title        = {Chasing Moving Targets with Online Self-Play Reinforcement Learning
                  for Safer Language Models},
  journal      = {CoRR},
  volume       = {abs/2506.07468},
  year         = {2025}
}

@article{zhang2025interplay,
  title={On the interplay of pre-training, mid-training, and rl on reasoning language models},
  author={Zhang, Charlie and Neubig, Graham and Yue, Xiang},
  journal={arXiv preprint arXiv:2512.07783},
  year={2025}
}

@article{SuperGPQA,
  author       = {M{-}A{-}P Team},
  title        = {SuperGPQA: Scaling {LLM} Evaluation across 285 Graduate Disciplines},
  journal      = {CoRR},
  volume       = {abs/2502.14739},
  year         = {2025}
}

@misc{aimo_amc_2023,
  author = {AI-MO},
  title  = {American Mathematics Contest},
  howpublished = {\url{https://huggingface.co/datasets/AI-MO/aimo-validation-amc}},
  year   = {2023}
}

\clearpage

\beginappendix
\section{Implementation Details} \label{apdx:impl-details}
We conduct the cold-start training with the LlamaFactory \footnote{https://github.com/hiyouga/LlamaFactory} and all the RL experiments based on the verl \footnote{https://github.com/volcengine/verl} framework. 

\textbf{Cold-start. }
For the cold-start stage, we collect 40k training data from the responses of Doubao-Seed-1.6-thinking \footnote{https://openrouter.ai/bytedance-seed/seed-1.6}, including 20k math reasoning trajectories and 20k rubricator trajectories, which are collected via reject sampling. 
We then perform full-parameter supervised fine-tuning (SFT) on the 40k dataset using Qwen3 prompt template, with a sequence cutoff length of 32768, per-device train batch size = 1, gradient accumulation steps = 1, base learning rate = 2e-5 (cosine scheduler with min lr = 2e-6), weight decay = 0.1, warmup ratio = 0.01 and total training epochs = 5.0. 
For the verifier, we distill a small-sized model Qwen3-4B-Base based on the verification responses from Doubao-Seed-1.6-thinking, using the same SFT recipe as in the cold-start stage.

\begin{table*}[h]
\centering
\begin{tabularx}{0.55\linewidth}{l|c}
  \toprule
\multicolumn{1}{c}{\textbf{Hyperparameter}} \vline & \multicolumn{1}{c}{\textbf{Value}} \\
\midrule
\small
Prompt template                       & Qwen3 \\
Sequence cutoff length                & 32768 \\
Per-device train batch size           & 1 \\
Gradient accumulation steps           & 1 \\
Learning rate (cosine scheduler)      & 2e-5 (min lr = 2e-6) \\
Weight decay                          & 0.1 \\
Warmup ratio                          & 0.01 \\
Epochs                                & 5.0 \\
\bottomrule
\end{tabularx}
\caption{Key training hyperparameters used in the cold-start SFT.}
\label{tab:sft_hyper}
\end{table*}

\clearpage
\textbf{RL training. }
For the RL training, we train all the models with the DAPO-Math-17K data, with prompt max length = 16384, response max length = 12288 (overlong buffer enabled, length = 4096), no KL reward/loss, clip ratio = 0.2, learning rate for actor = 1e-6, learning rate for critic = 1e-5, temperature = 1.0, top\_p = 1.0, train batch size = 32, rollout number $N$ = 8, mini batch size = 64, weight decay=0.1, grad clip=1.0, lr warmup steps=10. 
For all RL training, we fix the max training steps to 1500.
The response length and prompt length are configured in this setting because the outputs generated by the reasoner will serve as the input prompts for the rubricator. Therefore, the max prompt length should be greater than the response length, and the sum of these two values must not exceed 32k tokens.

\begin{table*}[h]
\centering
\begin{tabularx}{0.4\linewidth}{l|c}
  \toprule
\multicolumn{1}{c}{\textbf{Hyperparameter}} \vline & \multicolumn{1}{c}{\textbf{Value}} \\
\midrule
\small
Algorithm                             & PPO \\
Training dataset                      & DAPO-Math-17K \\
Max prompt length                     & 16384 \\
Max response length                   & 12288 \\
Overlong buffer                       & 4096 \\
KL reward/loss                        & None \\
Clip ratio                            & 0.2 \\
Learning rate (actor)                 & 1e-6 \\
Learning rate (critic)                & 1e-5 \\
Sampling temperature                  & 1.0 \\
Top\_p                                & 1.0 \\
Train batch size                      & 32 \\
Rollout number ($N$)                  & 8 \\
Mini batch size                       & 64 \\
Weight decay                          & 0.1 \\
Gradient clip                         & 1.0 \\
LR warmup steps                       & 10 \\
Max steps                             & 1500 \\
\bottomrule
\end{tabularx}
\caption{Key training hyperparameters used in RL training.}
\label{tab:rl_hyper}
\end{table*}

\clearpage
\section{Prompts} \label{apdx:prompts}
In this section, we report the prompts used as the reasoner $\pi_{\theta}^{\text{\textcolor{reasoner}{$Rea$}}}$ and the rubricator $\pi_{\theta}^{\text{\textcolor{rubricator}{$Rub$}}}$. It is worth mentioning that during our initial attempts to directly prompt Doubao-Seed-1.6-thinking model to generate rubrics based on the question and a reference answer, the model was unable to generate rubrics that are not already presented in the reference. Therefore, we modified the prompt to instruct the model to generate a set of rubrics where the current response score falls below the middle value, and explicitly reinforced this in the format reward. We observed that this practice triggers the model to actively refine rubrics during reasoning to generate meaningful rubrics for room of self-improvement. The 20k rubricator trajectories were then collected using this specific prompt with rejection sampling under such format reward.

\begin{figure}[!ht]
    \centering
    \begin{lrbox}{\DSRPrompt}
    \begin{gptpromptbox}
    \begin{minipage}{0.95\linewidth}
    \raggedright
    \scriptsize\ttfamily
\setlength{\baselineskip}{1\baselineskip}
\{question\}

Please reason step by step, and put your final answer within \textbackslash
 boxed\{\}.
\end{minipage}
    \end{gptpromptbox}
    \end{lrbox}
    \usebox{\DSRPrompt}
    \caption{The prompt of the reasoner role.}
    \label{fig:evaluation-prompts}
\end{figure}

\begin{figure}[ht]
    \centering
    \begin{lrbox}{\DSRPrompt}
    \begin{gptpromptbox}
    \begin{minipage}{0.95\linewidth}
    \raggedright
    \scriptsize\ttfamily
\setlength{\baselineskip}{1\baselineskip}
You are an expert in educational assessment and rubric design.
Your task is to analyze a given question–answer pair and generate comprehensive evaluation rubrics that can be used to assess response quality for this question. The answer is only a reference answer from the student(not necessarily a good response), so your rubric system should consider the merits already presented in the response, and most importantly, room for further improvements.

\# Input Data

[Question]:
\{question\}

[Response]:
\{response\}

\# Task Instructions

Based on the provided question and answer, generate a comprehensive rubric that's suitable with multiple evaluation criteria. Each criterion should be:\\
1. Specific and Measurable: Clearly define what constitutes meeting or not meeting the criterion\\
2. Binary Evaluable: Can be assessed as true/false by an LLM evaluator\\
3. Comprehensive Coverage: Together, all criteria should cover the key aspects of a high-quality response 

\# Required Rubric Categories

Generate criteria to cover these aspects:
1. Effectiveness of Problem Decomposition \& Planning
Definition: The systematic breakdown of a complex problem into logically ordered, manageable sub-tasks, and the formulation of a strategic roadmap for solving them.
Core Aspects: Identifying key components, establishing step dependencies, sequencing operations, and anticipating challenges before execution.

2. Effectiveness of BackTracking / Self-Validation / Error Handling
Definition: The continuous monitoring of reasoning validity through consistency checks, proactive error detection, and adaptive revision of flawed steps.
Core Aspects: Implementing verification checkpoints (e.g., unit analysis/estimation), diagnosing inconsistencies, recovering from dead ends via path correction (not restarting), and mitigating error propagation.

3. Reasoning Clarity \& Flow
Definition: The coherent structuring and communication of logical progression where each step explicitly follows from prior conclusions and leads to subsequent insights.
Core Aspects: Hierarchical argument organization under clear objective, unambiguous terminology, explicit logical connectors (e.g., "thus," "since," "conversely"), and seamless transitions between ideas.

4. Reasoning Focus \& Efficiency
Definition: The maintenance of strict alignment with problem objectives while optimizing cognitive resources through elimination of redundancies and irrelevant explorations.
Core Aspects: Sustained direction toward core goals, pruning of tangential paths, and avoidance of overthinking/repetition during reasoning.

5. Other Question-Specific Aspects
Definition: The strategic selection and application of domain-relevant methods that align with both problem constraints and solution goals.
Core Aspects: Leveraging context-optimal approaches (e.g., dimensional analysis in physics, elimination methods for multiple-choice), avoiding technique mismatches, and adapting tools to exploit problem structure (e.g., symmetry in geometry).

\# Output Format
Return a JSON object with the following structure:

```json\\
\{\\
  \quad"question\_domain": "math/calculus",\\
  \quad "rubrics": [
    \{
      "category": "Effectiveness of Problem Decomposition \& Planning" 
      "criterion": "XXX",
      "points": 5
    \},
    \{
      "category": "Reasoning Clarity \& Flow" 
      "criterion": "XXX",
      "points": 3
    \},
    \{
      "category": "Reasoning Focus \& Efficiency" 
      "criterion": "XXX",
      "points": 4
    \},
    \{
      "category": "Other Question-Specific Aspects" 
      "criterion": "XXX",
      "points": -3
    \},
    \{
      "category": "Reasoning Focus \& Efficiency" 
      "criterion": "XXX",
      "points": -2
    \}
    // ... additional criteria
  ],\\
  \quad "maximum\_score": a(sum of all positive metrics),\\
  \quad "minimum\_score": b(sum of all negative metrics),\\
  \quad "current\_score": x(must be smaller than (a+b)/2, otherwise revise your rubric system to get better room of improvement) \\
\}\\
```

\# Important Guidelines 
- First output the domain of the question in two levels, then generate 5 - 15 criteria in total, ensuring comprehensive coverage - Points should reflect the relative importance of each criterion (supports positive scores from 1 to 5 for reward criteria, and negative scores from -5 to -1 for penalty criteria). Specifically, a penalty criteria should be a statement that expresses flaw/failure in the reasoning process, while a reward criteria should be a statement that express merits/goal accomplishment in the reasoning process. Some of the final things that your should CHECK BEFORE YOUR FINAL RESPONSE:

\quad - Try to make your rubrics specific to the question, but not so detailed as some of the overly detailed metrics may not apply to all the response(e.g. "fail to apply XXX theorem" would be a acceptable metric, but "fail to reflect after miscalculating using XXX theorem in step25" would be a very bad metric to avoid, since another students response may not apply XXX theorem, nor having step 25). In other words, do not let the reference answer constrain your rubric system!  Again, DO NOT let the reference answer constrain your rubric system!!! To do this, try to propose a set of metrics pretending the reference answer is not there, and reconsider adding some of them. 

\quad - Remember, the answer above is only a reference answer from the student(not necessarily a good response), so your rubric system should consider the merits already presented in the response, and most importantly, room for further improvements. To enforce this, the reference response should have a score strictly below the average of your proposed rubric system. In order to check for this, you need to calculate the maximum\&minimum score from you rubric system, and score the given response at the end of your json. If you found that the rubrics you presented the first time gives a overly-high score, brainstorm more rubrics pretending the reference answer is not there in orderly to find room for improvements and add them to your rubric systems.
- Return only the JSON object without additional commentary.
    
    \end{minipage}
    \end{gptpromptbox}
    \end{lrbox}
    \usebox{\DSRPrompt}
    \caption{The prompt of the rubricator role.}
    \label{fig:evaluation-prompts}
\end{figure}

\begin{figure}[ht]
    \centering
    \begin{lrbox}{\DSRPrompt}
    \begin{gptpromptbox}
    \begin{minipage}{0.95\linewidth}
    \raggedright
    \scriptsize\ttfamily
\setlength{\baselineskip}{1\baselineskip}
You are an expert in evaluating student responses to math problems. Your task is to assess a given response based on a set of binary rubrics and compute a final score.

[Task Input]:

- Question: The math problem the student is solving.

- Response: The student's step-by-step solution.

- Rubrics: A list of criteria for evaluation. Each criterion has:

\quad - category: The aspect of reasoning being assessed.
    
\quad - criterion: A binary statement (either a merit to reward or a flaw to penalize).
    
\quad - points: The points assigned (positive for rewards, negative for penalties).

[Task Instruction]:\\
Evaluate each criterion:\\
- For each criterion in the rubrics list (in the given order), determine if the criterion is satisfied by the response.\\
\quad - Output True if the criterion is met (for a reward) or if the flaw is present (for a penalty).\\
\quad - Output False otherwise. \\
- Base your judgement solely on the content of the response and the specific wording of the criterion.\\
- Compute the final score: Sum the points of every criterion that is evaluated as True (this includes both positive and negative points).

[Output Format]\\
Return a JSON object with the following structure:\\
```json\\
\{\\
  \quad "judgement": [boolean, boolean, ...],  // List of True/False for each criterion in the same order as the rubrics\\
  \quad "final\_score": number                  // Sum of points for True judgements\\
\}\\
```

[Important Guidelines]

- Be objective: Do not consider any external knowledge beyond the provided question, response, and rubrics.

- Binary decision: Each criterion must be evaluated as strictly True or False. There are no partial credits.

- Order matters: The judgement list must exactly match the order of the rubrics provided.

- Independent evaluation: Assess each criterion separately. The presence or absence of one flaw/merit does not influence the evaluation of others (unless the criterion explicitly references another).

\quad- Penalty criteria: For a criterion with negative points, True means the flaw is present (so the negative points are added). False means the flaw is absent (so 0 is added for that criterion).

\quad- Reward criteria: For a criterion with positive points, True means the merit is present (so positive points are added). False means the merit is absent (so 0 is added).

Now, proceed to evaluate the following:

Question:
\{question\}

Response:
\{resp\}

Rubrics:
\{rubrics\}

Your final output should contain only the JSON object. Do not include any additional text or explanations.

\end{minipage}
    \end{gptpromptbox}
    \end{lrbox}
    \usebox{\DSRPrompt}
    \caption{The prompt of the verifier.}
    \label{fig:evaluation-prompts}
\end{figure}

\clearpage
\section{Case Study} \label{apdx:case-study}
In this section, we present a case to illustrate how the rubricator proposes meaningful metrics conditioned on the reasoner's response. 
This example shows that the rubricator successfully identifies the key problem(\textit{Employs ad-hoc manual listing for small digit sets instead of systematic permutation formula application, increasing error risk}) that caused miscalculation for $d=0$ and $d=1$ subsets, leading to the final error of 131. It also gives a hint to efficiently solving the problem and reducing reasoning redundancy (\textit{Avoids redundant recounting by categorizing sets with minimum permutation $>$ 2013 upfront, reducing per-set analysis effort}). Therefore, the self-proposed rubrics can effectively guide the model to solve the specific problem, serving as valuable auxiliary reward for the reasoner.

\textbf{Question:}
2013 is the first year since the Middle Ages that consists of four consecutive digits. How many such years are there still to come after 2013 and before 10000?

\textbf{Ground Truth:}
149

\textbf{Prediction:}
131

\textbf{Generated Rubrics:}
\begin{figure}[!ht]
    \centering
    \begin{lrbox}{\DSRPrompt}
    \begin{promptbox}
    \raggedright
    \scriptsize\ttfamily
\setlength{\baselineskip}{1\baselineskip}
\{\\
\hspace{2em}"question\_domain":"math/digit-sequence-analysis",\\
\hspace{2em}"rubrics": [\\
\hspace{4em}\{\\
\hspace{6em}"category":"Effectiveness of Problem Decomposition \& Planning",\\
\hspace{6em}"criterion":"Explicitly outlines digit set ranges, leading zero constraints, and year boundary limits before permutation calculations",\\
\hspace{6em}"points":3\\
\hspace{4em}\},\\
\hspace{4em}\{\\
\hspace{6em}"category":"Reasoning Clarity \& Flow",\\
\hspace{6em}"criterion":"Uses hierarchical structuring (e.g., bullet points) for each digit set with explicit permutation counts and exclusion justifications",\\
\hspace{6em}"points":3\\
\hspace{4em}\},\\
\hspace{4em}\{\\
\hspace{6em}"category":"Reasoning Focus \& Efficiency",\\
\hspace{6em}"criterion":"Avoids redundant recounting by categorizing sets with minimum permutation > 2013 upfront, reducing per-set analysis effort",\\
\hspace{6em}"points":3\\
\hspace{4em}\},\\
\hspace{4em}\{\\
\hspace{6em}"category":"Effectiveness of BackTracking, Self-Validation \& Error Handling",\\
\hspace{6em}"criterion":"Fails to implement edge case validation (e.g., overcounting leading-zero permutations) via secondary checklist reviews",\\
\hspace{6em}"points":-4\\
\hspace{4em}\},\\
\hspace{4em}\{\\
\hspace{6em}"category":"Other Question-Specific Aspects",\\
\hspace{6em}"criterion":"Employs ad-hoc manual listing for small digit sets instead of systematic permutation formula application, increasing error risk",\\
\hspace{6em}"points":-2\\
\hspace{4em}\}\\
\hspace{2em}],\\
\hspace{2em}"maximum\_score":9,\\
\hspace{2em}"minimum\_score":-6\\
\}
    \end{promptbox}
    \end{lrbox}
    \usebox{\DSRPrompt}
    \caption{The rubricator's generated rubrics}
    \label{fig:evaluation-prompts}
\end{figure}

\clearpage
\textbf{Reasoner's Response:}

\begin{figure}[!ht]
    \centering
    \begin{lrbox}{\DSRPrompt}
    \begin{promptbox}
    \raggedright
    \scriptsize\ttfamily
\setlength{\baselineskip}{1\baselineskip}

Got it, let's try to figure out this problem: we need to find how many years after 2013 and before 10000 have four consecutive digits. First, let's make sure I understand what "four consecutive digits" means in a year (which is a four-digit number, right? Wait, years like 2013 are four-digit, 10000 is five-digit). So a four-digit year with four consecutive digits—wait, does that mean the digits are consecutive in order, like 2013 is 2,0,1,3? Wait no, 2013 isn't consecutive digits. The problem says "2013 is the first year since the Middle Ages that consists of four consecutive digits"—wait, maybe it's digits that are consecutive integers, but not necessarily in order? Wait, wait, let's recheck: "consists of four consecutive digits"—probably, it means the set of digits is four consecutive integers, maybe in any order? Wait, 2013: digits are 2,0,1,3—are those four consecutive? 0,1,2,3 are four consecutive, but 2013 has 0, so 0 is there, then 1,2,3 but missing 2? Wait no, 0,1,2,3 are four consecutive numbers, but 2013 has digits 2,0,1,3—wait maybe I misread: 2013's digits: 2,0,1,3—wait 0,1,2,3 are four consecutive, but is 2013 a permutation of four consecutive digits? Let's list permuted four-digit numbers with digits being four consecutive integers.\\
First, what are four consecutive integers? Let's denote four consecutive digits as d, d+1, d+2, d+3 where d is from 0 to 6 (since d+3 $\leq$ 9, because digits go up to 9). So possible d: 0 (0-3), 1 (1-4), 2 (2-5), 3 (3-6), 4 (4-7), 5 (5-8), 6 (6-9). Wait let's list the sets:\\
d=0: \{0,1,2,3\}\\
d=1: \{1,2,3,4\}\\
d=2: \{2,3,4,5\}\\
d=3: \{3,4,5,6\}\\
d=4: \{4,5,6,7\}\\
d=5: \{5,6,7,8\}\\
d=6: \{6,7,8,9\}\\
These are the possible sets of four consecutive digits (digits are 0-9, so four consecutive can't include 7-10, etc.). Now, a four-digit year can't start with 0, so we need to count permutations of each set that are four-digit numbers (so first digit $\ne$ 0) and are between 2013 and 10000 (so starting from 2014 up to 9999). Wait 2013 is the first since Middle Ages—so we need years after 2013, so starting from 2014.\\
Let's take each set and count valid permutations (four-digit, no leading zero, >2013):\\
\hl{Set \{0,1,2,3\} (d=0): How many four-digit permutations? Total permutations: 4! = 24, minus those starting with 0: 3! = 6, so 18. Now we need permutations >2013. Let's list or calculate: First digit: can be 1,2,3 (since 0 is invalid). If first digit is 1: then the rest are permutations of 0,2,3: 3! = 6 numbers (1023, 1032, 1203, 1230, 1302, 1320). The smallest in this first digit group is 1023, which is less than 2013, so we can ignore all first digit 1 permutations. If first digit is 2: then permutations starting with 2: the remaining digits are 0,1,3. We need permutations >2013. The smallest permutation starting with 2 is 2013, which is the given year. So permutations starting with 2 and >2013: remaining digits 0,1,3—how many are >2013? The smallest is 2013 (which is equal), next would be 2031, 2103, 2130, 2301, 2310. Wait let's count: For first digit 2, second digit: can be 0,1,3 (since 2 is already used). If second digit is 0: then third and fourth digits are permutations of 1,3: 2 numbers (0132 invalid? Wait no, first digit 2, second 0: 2 0 x x—so numbers are 2013, 2031 (since 1 and 3 are digits, so 2013 and 2031). 2013 is equal, so 2031 is >2013. If second digit is 1: then third and fourth digits are permutations of 0,3: numbers are 2103, 2130 (both >2013). If second digit is 3: then third and fourth digits are permutations of 0,1: numbers are 2301, 2310 (both >2013). So for first digit 2: number of permutations >2013: let's count: 2031 (yes), 2103 (yes), 2130 (yes), 2301 (yes), 2310 (yes). That's 5 numbers from set \{0,1,2,3\}.
Set \{1,2,3,4\} (d=1): Four-digit numbers, no leading zero (so first digit 1-4). All permutations are valid since first digit can't be 0. Total permutations: 4! = 24. We need years after 2013. The smallest number in this set is 1234, which is less than 2013, so many numbers here are below 2013. Wait first digit 1: numbers start with 1, so 1xxx, which are all <2000, so ignore. First digit 2: permutations starting with 2: total 6 (since second digit: 3! for remaining 3 digits). These are from 2134 upwards? Wait list them: 2134, 2143, 2314, 2341, 2413, 2431. All these are >2013, right? Yes, since smallest is 2134, which is >2013. So all 6 permutations starting with 2 in this set are valid.}\textbf{==> Employs ad-hoc manual listing for small digit sets instead of systematic permutation formula application, increasing error risk}\\
\hl{Set \{2,3,4,5\} (d=2): First digit can be 2-5 (since no 0 here). So permutations: total 4! =24, minus leading zero? Wait no, no 0, so all 24 are valid four-digit (since first digit 2-5). We need years after 2013. Let's list first digits: 2,3,4,5. First digit 2: permutations starting with 2: 6 numbers (2 \_ \_ \_). The smallest is 2345? Wait no, digits are 2,3,4,5—wait permutations starting with 2: second digit can be 3,4,5 (since 2 is used). So numbers: 2345, 2354, 2435, 2453, 2534, 2543. The smallest in first digit 2 is 2345, which is greater than 2013. So all 6 permutations starting with 2 are valid. First digit 3: permutations starting with 3: 6 numbers (all >2345, so >2013), same for 4 and 5: all 6 each. So total for this set: 6 + 6 + 6 + 6 = 24.\\
Set \{3,4,5,6\} (d=3): Similar to above, digits 3-6, no 0/1. So all four-digit permutations are valid (first digit 3-6). All numbers here are $\geq$ 3456, which is way above 2013, so all 4! =24 permutations are valid.\\
Set \{4,5,6,7\} (d=4): Even higher, 4xxx and above, all permutations valid: 24 numbers.\\
Set \{5,6,7,8\} (d=5): 5xxx and above: 24 numbers.\\
Set \{6,7,8,9\} (d=6): 6xxx and above: 24 numbers.}\textbf{==> Avoids redundant recounting by categorizing sets with minimum permutation > 2013 upfront, reducing per-set analysis effort}\\
Now let's add them up:\\
Set \{0,1,2,3\}: 5\\
Set \{1,2,3,4\}: 6\\
Set \{2,3,4,5\}: 24\\
Set \{3,4,5,6\}: 24\\
Set \{4,5,6,7\}: 24\\
Set \{5,6,7,8\}: 24\\
Set \{6,7,8,9\}: 24\\
Calculations: 5 + 6 = 11; 11 + 24*5 = 11 + 120 = \boxed{\textbf{131}}.\\
    \end{promptbox}
    \end{lrbox}
    \usebox{\DSRPrompt}
    \caption{The reasoner's response}
    \label{fig:evaluation-prompts}
\end{figure}

\end{document}